\documentclass[sigconf]{acmart}
\usepackage{graphicx}  

\usepackage[linesnumbered,ruled,vlined]{algorithm2e}
\usepackage{pbox}
\usepackage{amsmath}
\usepackage{amsthm}
\usepackage{subcaption}
\usepackage{thmtools, thm-restate}
\usepackage{multirow}
\usepackage{mathrsfs}
\usepackage{bbm}
\usepackage{colortbl}
\usepackage{commath}
\usepackage{centernot}
\usepackage{amssymb}
\usepackage{mathtools}
\usepackage{enumitem}

\def\BibTeX{{\rm B\kern-.05em{\sc i\kern-.025em b}\kern-.08emT\kern-.1667em\lower.7ex\hbox{E}\kern-.125emX}}
    
\newtheorem{proposition}{Proposition}
\newtheorem{definition}{Definition}
\declaretheorem{theorem}

\begin{document}

%
\title{Crowdsourcing with Fairness, Diversity and Budget Constraints}

%
\author{Naman Goel}
\email{naman.goel@epfl.ch}
\affiliation{%
  \institution{	Artificial Intelligence Lab, EPFL\\
  	Lausanne, Switzerland\\}
}

\author{Boi Faltings}
\email{boi.faltings@epfl.ch}
\affiliation{%
	\institution{	Artificial Intelligence Lab, EPFL\\
		Lausanne, Switzerland\\}
}

\copyrightyear{2019} 
\acmYear{2019} 
\setcopyright{acmcopyright}
\acmConference[AIES '19]{AAAI/ACM Conference on AI, Ethics, and Society}{January 27--28, 2019}{Honolulu, HI, USA}
\acmBooktitle{AAAI/ACM Conference on AI, Ethics, and Society (AIES '19), January 27--28, 2019, Honolulu, HI, USA}
\acmPrice{15.00}
\acmDOI{10.1145/3306618.3314282}
\acmISBN{978-1-4503-6324-2/19/01}

%

%
\begin{abstract}
Recent studies have shown that the labels collected from crowdworkers can be discriminatory with respect to sensitive attributes such as gender and race. This raises questions about the suitability of using crowdsourced data for further use, such as for training machine learning algorithms. In this work, we address the problem of fair and diverse data collection from a crowd under budget constraints. We propose a novel algorithm which maximizes the expected accuracy of the collected data, while ensuring that the errors satisfy desired notions of fairness. We provide guarantees on the performance of our algorithm and show that the algorithm performs well in practice through experiments on a real dataset.
\end{abstract}

%
%


%

%
\maketitle
\section{Introduction}
Algorithmic decision-making is gaining popularity in many diverse application areas of social importance. Examples include criminal recidivism prediction, stop-and-frisk programs, university admissions, bank loan decisions, screening job candidates, fake news control, information filtering(personalization) and search engine rankings etc. Recently, questions were raised about the fairness of these algorithms. An investigation~\cite{pro-publica} found COMPAS (a popular software used by courts to predict criminal recidivism risk) racially discriminatory. Other software systems have also been found to be biased against people of different races, genders and political views~\cite{kay2015unequal,bolukbasi2016man,otterbacher2017competent,kulshrestha2017quantifying}. This has led to a widespread and legitimate concern about the potential negative influence of such systems on the society~\cite{barocas2016big,president-statement}. One of the main reasons of algorithmic bias is the bias in the training datasets. In order to achieve algorithmic fairness, the issue of \textbf{data fairness} needs to be addressed first. In many interesting cases, data is directly or indirectly influenced by some kind of human feedback. The influence is obvious and direct if human assigned labels are used as a proxy for ground truth labels. However, human feedback can also indirectly influence the so-called ``ground-truth" datasets (when the labels are not human assigned but observed in reality). This is because the ground truth labels can only be collected for a finite number of data points and the selection of data points is often influenced by humans. For example, there are no ground truth labels available for recidivism of people who were never released by the judges. In this paper, we focus only on the direct influence of human feedback on data fairness i.e. the case in which humans assign labels for data.

Crowdsourcing is increasingly used to collect training data labels. Inevitably, crowdworkers have different biases, which are then reflected in the labels collected from the workers. A very recent study~\cite{dressel2018accuracy} conducted on Amazon Mechanical Turk showed that the crowdworkers were equally racially biased as COMPAS in predicting recidivism. The difference in false positive rates of crowd predictions for white and black defendants was significant and nearly equal to that of the predictions made by COMPAS. The same was true for false negative rates also. The bias didn't change much even when the crowdworkers were not explicitly displayed the race of the defendants.

We consider settings similar to~\cite{dressel2018accuracy}. Workers are asked to provide their answers (or labels) about some tasks with unknown ground truth labels. Every task has some non-sensitive details that are shown to the workers and a sensitive attribute (for example, race) that is not explicitly shown. But the sensitive attribute may potentially be correlated with the non-sensitive task details. A worker inspects the tasks assigned to her and submits labels for the tasks. Each task is assumed to have a ground truth label but the workers don't have any way of accessing the ground truth. They can only use the task details, their prior knowledge and incomplete information from other sources to make an ``educated guess" about the ground truth. The examples of such tasks are ``Will a defendant with given personal history recidivate within the next two years or not?" or ``Will a candidate with given CV be successful in the job applied for?" or ``Is given political news item fake?". The sensitive attributes in these example tasks are race, gender and political group respectively. Every worker charges a fee for answering the assigned tasks. The requester has a budget constraint on the fees that she can pay to the workers. In this paper, we make the following contributions:
\begin{enumerate}
	\item We propose a novel algorithm for assigning tasks to workers, which optimizes the expected accuracy of labels obtained from crowd while ensuring that the collected labels satisfy desired notions of \textit{error fairness}. The algorithm also ensures diversity of responses by limiting the probability of assigning many tasks to a single worker. Our algorithm works even when the values of the sensitive attribute of the tasks are unavailable or can't used because of ethical/legal reasons.
	\item With a novel formulation of the task assignment strategy as a probability distribution over the workers, we can cast the optimization problem as a linear program and avoid the use of integer programming or other graph matching algorithms which are popular in the task assignment literature but are harder to solve exactly and analyze. This also makes our algorithm suitable for online settings in which the requester is not aware of the tasks in advance.
	\item We use a limited number of gold tasks (tasks with known ground-truth answers) for estimating workers' parameters and then optimally assign non-gold tasks to the workers. We provide performance bounds for our algorithm and show empirical performance on a real dataset. \end{enumerate}

\section{Related Work}

\textbf{Empirical Studies : } \cite{dressel2018accuracy} finds racial discrimination in recidivism prediction tasks on Amazon Mechanical Turk (AMT). \cite{otterbacher2015crowdsourcing} analyzes linguistic bias in labels collected through GWAP (Games with a Purpose) on AMT. \cite{otterbacher2015linguistic} analyzes the linguistic bias in collaboratively produced biographies. \cite{hannak2017bias} finds discrimination in reputation crowdsourcing systems in online marketplaces.

\noindent \textbf{Proposed Solutions : }In an independent and pioneering work,~\cite{valera2018enhancing} considers the problem of fairness in human decision-making tasks like recidivism prediction, without budget and diversity constraints. Criminal cases with \textit{known} race information arrive in batches of \textit{known} sizes and an MDP based maximum weighted matching algorithm assigns each case to \textit{exactly one} human judge such that the overall utility from decisions of releasing or keeping any defendant is maximized, while ensuring \textit{demographic parity} of release decisions across two races. To the best of our knowledge this is the first and the most recent work to consider settings somewhat similar to ours but our work differs from theirs in several ways. We consider general crowdsourcing settings, in which several assumptions from their model don't hold. In particular, they assume that ``true" risk scores of individual defendants are known to the human judges and the case assignment algorithm. In general crowdsourcing settings, one can only hope to have an overall label distribution for the population. In fact, finding the label probability for individual tasks is the very objective of crowdsourcing. Further, it is not immediately clear how their work can be extended for other important fairness definitions. In their model, given true risk scores of the defendants, judges only apply different thresholds for black and white defendants to predict recidivism. The threshold parameters alone can't capture unfairness measures such as unequal error rates. Even if one does improvise the model with more parameters, it remains an open question whether the theoretical conjectures made in the paper are still likely. This is because the conjectures assume that every time a judge gives a decision, the model parameters of the judge are updated. This becomes an issue with error rate parameters since the ground truth labels are not revealed for all tasks in crowdsourcing. \cite{sethneel} considers a different but related problem of bias resulting from adaptive data gathering (when the choice of whether to collect more data of a given type depends on the data already collected) and propose a differentially private data collection process as a solution. 

There is also a lot of work on task assignment in crowdsourcing, which doesn't consider fairness. \cite{tran2014efficient} proposes a greedy knapsack approach to satisfy limits on budget and the number of tasks any worker can solve. \cite{Karger:2014:BTA:2765242.2765244,ho2012online,ho2013adaptive} consider task assignment problem when workers arrive online. \cite{bragg2016optimal} proposes optimal gold task assignment when workers' diligence change over time.

Beyond data collection, there is also recent work on making algorithms fair and robust to bias in the training data~\cite{dwork2012fairness,hardt2016equality,zafar2017fairness-www,goel2018non,Kleinberg2017InherentTI} and on correcting bias in training datasets \cite{feldman2015certifying,calmon2017optimized}. Correcting bias in a given dataset requires modifying the feature values and/or the labels in the dataset. In this paper, we aim to collect unbiased and high quality dataset to begin with, relaxing the responsibility and the overhead of such post-processing from data users (for example, data scientists and machine learning engineers).


\section{Model}
Let there be a finite set of $n$ workers and a large pool of tasks with unknown ground truth labels. The data requester randomly chooses tasks from the pool one by one and assigns each to one (or more) worker(s). The requester may not have knowledge of all the tasks in the pool (not even the number of tasks in the pool) in advance. A worker $i$ charges a constant amount of fee $c_i$ for every label she provides. The requester has a budget constraint for the maximum \textit{expected} money to be spent on acquiring one label from a worker.

Let $Z$ be a random variable denoting the sensitive attribute and $Y$ denoting the (unknown) ground truth labels of the tasks such that $Z, Y \in \{0,1\}$. For the tasks attempted by a worker $i$, let $\hat{Y}_{i}\in \{0,1\}$ denote the labels submitted by the worker. We denote the realizations of random variables $Z$, $Y$ and $\hat{Y}_{i}$ by lower case letters $z$, $y$ and $\hat{y}_{i}$ respectively and will drop the subscripts for brevity when the context is clear. We will use $[n]$ to denote $\{1,2,\ldots,n\}$. The workers are modeled using their accuracy matrices as follows:
\begin{definition}[Accuracy Matrices of a Worker]
	The accuracy matrices $\mathcal{A}_{iz},\  z\in\{0,1\}$ of a worker $i$ are two $2 \times 2$ row stochastic matrices such that, $\forall y,\hat{y_i} \in \{0,1\}$, the entry $\mathcal{A}_{iz}[y,\hat{y_i}]$ is the probability of the worker's label on a task being $\hat{y_i}$ given that the sensitive attribute of the task is $z$ and the ground truth label is $y$.
\end{definition}
The two matrices $\mathcal{A}_{i0}$ and $\mathcal{A}_{i1}$ define the accuracy of the worker $i$ for tasks belonging to the two different values of the sensitive attribute. The accuracy matrix model, also known as the Dawid-Skene model~\cite{dawid1979maximum} in the crowdsourcing literature, is strong enough to capture different errors (for e.g. false positive and false negative rates) that a worker may make for tasks belonging to a given sensitive attribute value. If a worker is unbiased in the sense that her errors don't depend on the value of sensitive attribute of the task, her two accuracy matrices are identical. Note that this model makes an implicit i.i.d. assumption on a worker's answers.

The requester uses a probabilistic policy to assign the tasks to workers and collects the labels from the workers.
\begin{definition}[Crowdsourcing Policy]\label{def:policy}
	A crowdsourcing policy is an $n$-dimensional stochastic vector $S$, such that an element $S[i], i \in [n]$ is the probability of assigning any task to worker $i$, regardless of the sensitive attribute value of the task.
\end{definition}
Note that the requester's policy doesn't depend on the value of the sensitive attribute of the task. This is an intentional modeling choice to deal with the situations in which the sensitive attribute values of the tasks may not be available. It may be due to missing data, privacy reasons or legal/ethical requirements of not using the sensitive attribute.

For any task, the requester randomly selects one (or more than one) worker(s) with probabilities specified by the crowdsourcing policy vector $S$ and assigns the task to the selected worker(s). The labels collected from the workers are obviously not guaranteed to be error free. We can define the accuracy matrices of the crowdsourcing policy in the same way as we defined the accuracy matrices of workers.

\begin{definition}[Accuracy Matrices of a Crowdsourcing Policy]
	The accuracy matrices $\mathcal{A}_{z},\  z\in\{0,1\}$ of a crowdsourcing policy are two $2 \times 2$ row stochastic matrices such that, $\forall y,\hat{y} \in \{0,1\}$, the entry $\mathcal{A}_{z}[y,\hat{y}]$ is the probability that a crowdsourced label\footnote{We note that the accuracy of a crowdsourcing policy can also be defined in terms of aggregated label when multiple labels per task are collected. But such definitions depend on specific label aggregation algorithms used. However, in many cases, it may be sufficient to assume that the accuracy of a policy with aggregated labels is an increasing function of our accuracy, which is a reasonable assumption.} for a task is $\hat{y}$ given that the sensitive attribute of the task is $z$ and the ground truth label is $y$.
\end{definition}
We use the letter $\mathcal{A}$ to denote accuracy matrices of the crowdsourcing policy as well as that of the workers but readers can differentiate between the two by noting that $\mathcal{A}$ has an additional subscript $i$ when referring to the matrix of a worker $i$. It is easy to see that we can express the accuracy matrices of a policy in terms of the accuracy matrices of the workers as follows:
\begin{equation}\label{eq:convex}
	\mathcal{A}_z = \sum_{i = 1}^{n} S[i] \cdot \mathcal{A}_{iz} \qquad ,\forall z \in \{0,1\}
\end{equation}
The requester is interested in finding a crowdsourcing policy that maximizes the expected accuracy of the collected labels while ensuring that the data is \textit{fair}, \textit{diverse} and is acquired within budget constraints.

Crowd diversity is a subjective property and is generally defined in terms of the demographics of crowdworkers. In this paper, we work with a given set of crowdworkers and can't control such a measure of diversity. For settings like these, we define diversity as follows:
\begin{definition}[$\beta$-Diverse Crowdsourcing Policy]\label{def:diverse}
	A crowdsourcing policy is called $\beta$-diverse if and only if $\forall\ i \in [n], S[i]$ is upper bounded by $\beta$, where $\beta$ is a diversity parameter such that $0 \leq \beta <  1$.
\end{definition}
This definition limits the influence of individual workers on the overall crowdsourced dataset and aims to distribute the influence across more workers.

Similar to diversity, fairness is also a subjective property. We use some standard definitions of fairness from the machine learning literature~\cite{hardt2016equality,zafar2017fairness-www,fairness-book}. 
\begin{definition}[False Positive Rate Parity]\label{def:fpr}
	A crowdsourcing policy, with accuracy matrices $\mathcal{A}_0$ and $\mathcal{A}_1$, is said to satisfy false positive rate parity if and only
	$$\mathcal{A}_0[0,1] = \mathcal{A}_1[0,1]$$
\end{definition}
One can similarly define false negative rate parity, which requires $\mathcal{A}_0[1,0] = \mathcal{A}_1[1,0]$.
\begin{definition}[Error Rate Parity]\label{def:eqerror}
	A crowdsourcing policy, with accuracy matrices $\mathcal{A}_0$ and $\mathcal{A}_1$, is said to satisfy error rate parity if and only if it satisfies both false positive rate parity and false negative rate parity, i.e. $$\mathcal{A}_0 = \mathcal{A}_1$$
\end{definition}
It is easy to see that if all workers are unbiased, any crowdsourcing policy satisfies the above fairness definitions and one only needs to select a policy that maximizes accuracy while satisfying budget and diversity constraints. In this paper, we address the general problem scenario (when workers are not necessarily unbiased).
\section{Finding Optimal Crowdsourcing Policy}
Let's first assume that the accuracy matrices of all the workers are known and the requester is interested in finding the optimal crowdsourcing policy maximizing the expected accuracy under budget, fairness and diversity constraints. We model this as a constrained optimization problem. The objective function in the minimization problem is the negative of the expected accuracy of the policy variable $S$:
\begin{small}
	\begin{equation}\label{eq:expected-utility-2}
		\begin{split}
			-\mathbb{E}[\mathcal{A}&(S)] = -\kern-0.5em\sum_{z \in \{0,1\}} \kern-0.5em P(Z= z)\kern-0.5em\sum_{y \in \{0,1\}} \kern-0.5emP_z(Y=y) \sum_{i = 1}^{n} S[i] \mathcal{A}_{iz}[y,y]
		\end{split}
	\end{equation}
\end{small}where $P(Z=z)$ is the known prior probability that any random task in the pool will have sensitive attribute value equal to $z$ and $P_z(Y=y)$ is the known prior probability that any random task with sensitive attribute value $z$ in the pool will have a ground truth label equal to $y$.

Together with the fairness and diversity constraints, we get the following optimization problem:

\begin{small}
	\begin{equation}\label{eq:optimization-2}
		\begin{aligned}
			& \underset{S}{\text{arg min}}
			& & \kern-0.7em-\kern-0.7em\sum_{z \in \{0,1\}} \kern-0.7em P(Z= z)\kern-0.7em\sum_{y \in \{0,1\}} \kern-0.7emP_z(Y=y) \sum_{i = 1}^{n} S[i] \mathcal{A}_{iz}[y,y] \\
			& \text{subject to}
			& & \sum_{i = 1}^{n} S[i] = 1 \\
			& & & S[i] \geq 0\quad,\forall i \in [n]\\
			& & & S[i] \leq \beta\quad,\forall i \in [n]\\
			& & & \mathcal{A}_0[0,1] - \mathcal{A}_1[0,1] \leq \alpha\\
			& & & -(\mathcal{A}_0[0,1] - \mathcal{A}_1[0,1]) \leq \alpha\\
			& & & \sum_{i = 1}^{n} S[i] \cdot c_i \leq C
		\end{aligned}
	\end{equation}
\end{small}
The first two constraints are due to the fact that the crowdsourcing policy vectors are probabilistic and so, all elements must be positive and sum to $1$. The third is the diversity constraint as formalized in Definition~\ref{def:diverse}. The forth and fifth constraints together are equivalent to $\abs{\mathcal{A}_0[0,1] - \mathcal{A}_1[0,1]} \leq \alpha$. For $\alpha = 0$, we get the exact fairness constraint (false positive rate parity) as formalized in Definition~\ref{def:fpr}. Other fairness constraints can also be similarly included. The last constraint is due to the maximum expected budget ($C$) that can be spent on acquiring one answer from a worker.

\subsection{Estimates of Worker Accuracy Matrices}\label{sec:unknown-matrices}
Until now, we assumed that the accuracy matrices of the workers are known. However, in practice, we need to estimate them. As is common in the literature~\cite{oleson2011}, we assume that the requester has some limited number of gold standard tasks. Gold tasks are the tasks for which the requester not only knows the sensitive attribute value $z$ but also the ground truth label $y$. We use gold tasks to estimate unknown worker accuracy matrices. Estimating all the entries of the worker accuracy matrices requires that every worker answers some gold tasks of each ``type" (the type of a task is specified by its ground truth answer and its sensitive attribute value). We assign $N_{g}$ tasks of every type to each worker to estimate their accuracy matrices. This is explained further in appendix A. Let $\hat{\mathcal{A}}_{iz}$ be the estimate of the worker accuracy matrices $\mathcal{A}_{iz},\  \forall z\in\{0,1\}$. The optimization problem~\ref{eq:optimization-2} can now be written as follows, by replacing the accuracy matrices with their estimates:
\begin{small}
	\begin{equation}\label{eq:optimization-3}
		\begin{aligned}
			& \underset{S}{\text{arg min}}
			& & \kern-0.7em-\kern-0.7em\sum_{z \in \{0,1\}} \kern-0.7em P(Z= z)\kern-0.7em\sum_{y \in \{0,1\}} \kern-0.7emP_z(Y=y) \sum_{i = 1}^{n} S[i] \hat{\mathcal{A}}_{iz}[y,y] \\
			& \text{subject to}
			& & \sum_{i = 1}^{n} S[i] = 1 \\
			& & & S[i] \geq 0\quad,\forall i \in [n]\\
			& & & S[i] \leq \beta\quad,\forall i \in [n]\\
			& & & \hat{\mathcal{A}}_0[0,1] - \hat{\mathcal{A}}_1[0,1] \leq \alpha\\
			& & & -(\hat{\mathcal{A}}_0[0,1] - \hat{\mathcal{A}}_1[0,1]) \leq \alpha\\
			& & & \sum_{i = 1}^{n} S[i] \cdot c_i \leq C
		\end{aligned}
	\end{equation}
\end{small}
where,
\begin{small}
	\begin{equation}\label{eq:convex-3}
		\hat{\mathcal{A}}_z = \sum_{i = 1}^{n} S[i] \cdot \hat{\mathcal{A}}_{iz} \qquad ,\forall z \in \{0,1\}
	\end{equation}
\end{small}
This is a linear program, which can be exactly solved in polynomial time. In practice, the simplex method~\cite{chvatal1983linear} can be used to find the optimal solution efficiently with common optimization libraries like IBM CPLEX and SciPy. Depending on the constraints, the cost and the accuracy matrices of workers, it is possible that no feasible solution exists for the optimization problem. In this case, the requester will have no choice but to relax the constraints. 

We will now analyze our algorithm theoretically and empirically. Readers can find a summary of steps of our complete crowdsourcing algorithm in appendix B.
\begin{figure*}
	\centering
	\begin{subfigure}{0.33\textwidth}
		\centering
		\includegraphics[width=1\textwidth, trim={20 20 14 20},clip]{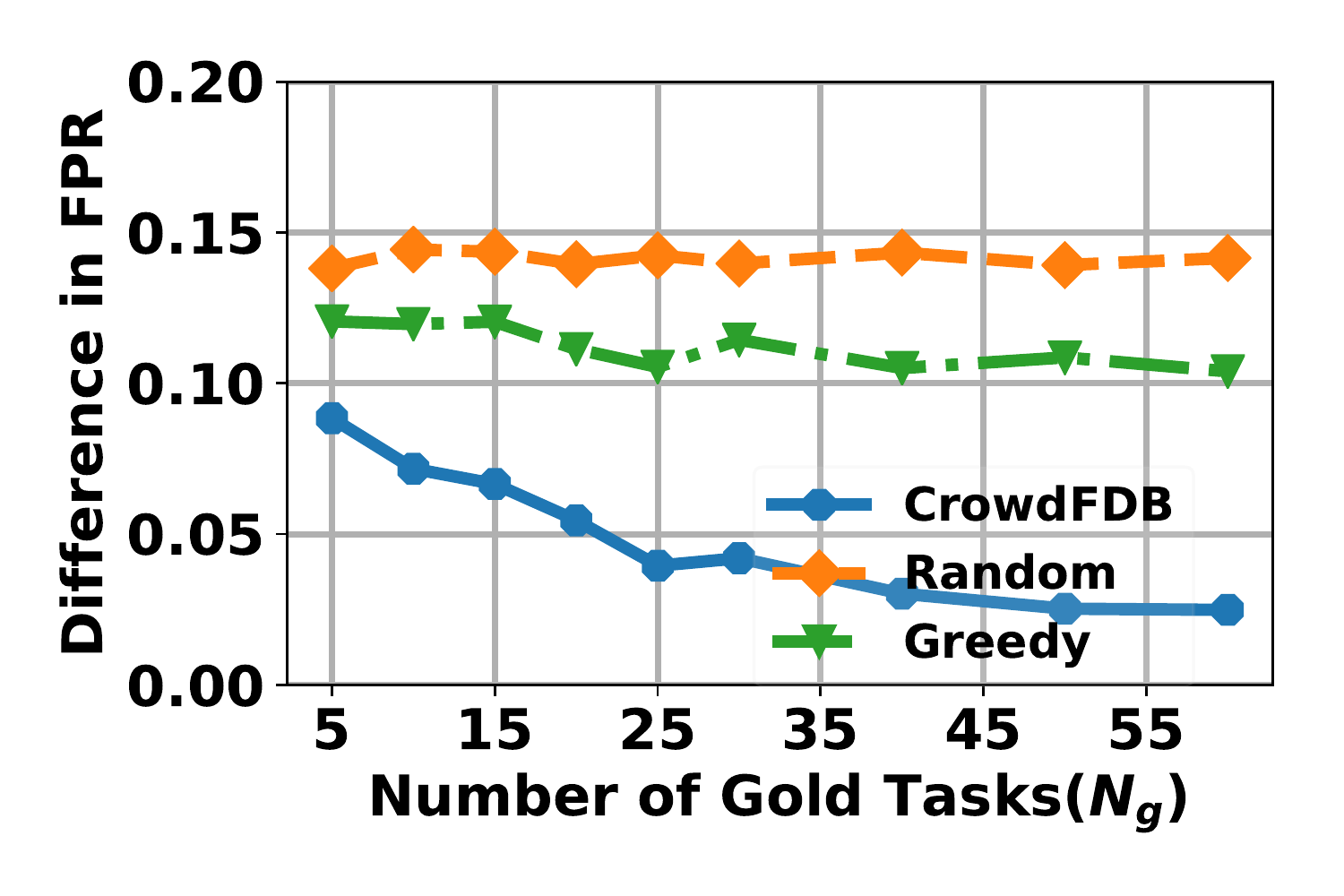}
		\caption{Absolute Difference in FPR}
		\label{fig:fpr-1} 
	\end{subfigure}
	\begin{subfigure}{0.33\textwidth}
		\centering
		\includegraphics[width=1\textwidth, trim={20 20 14 20},clip]{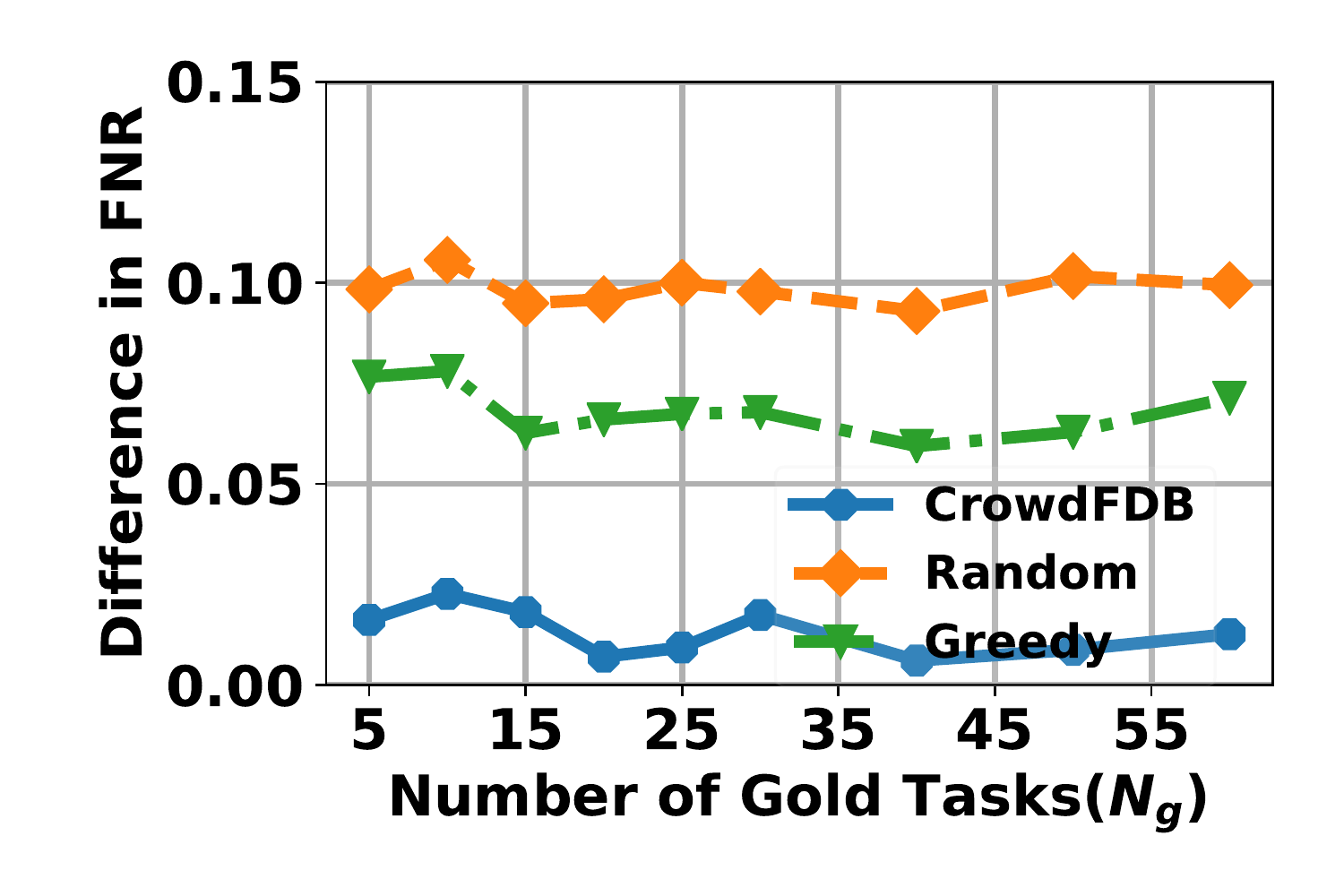}
		\caption{Absolute Difference in FNR}
		\label{fig:fnr-1} 
	\end{subfigure}
	\begin{subfigure}{0.33\textwidth}
		\centering
		\includegraphics[width=1\textwidth, trim={20 20 14 20},clip]{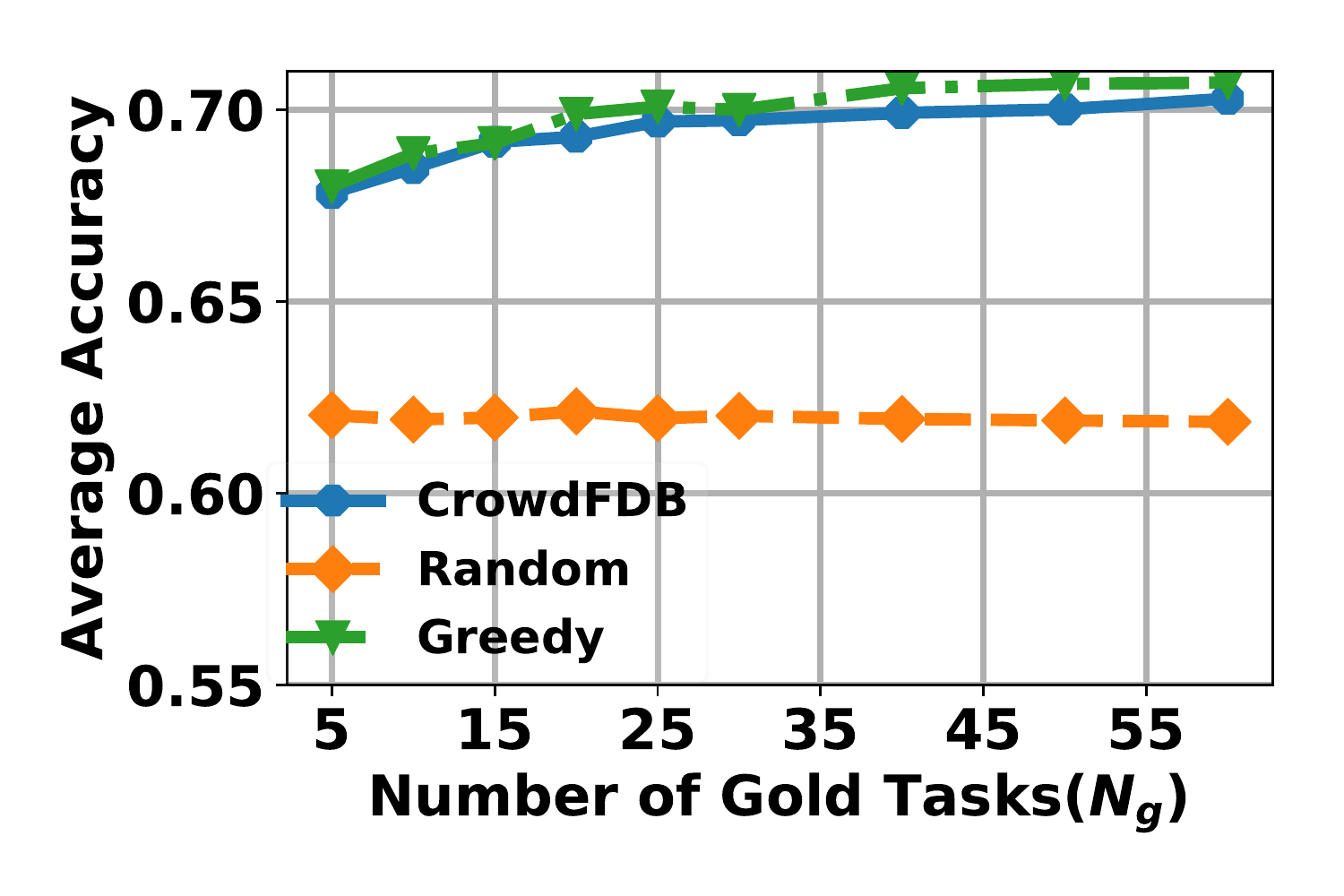}
		\caption{Average Accuracy}
		\label{fig:acc-1} 
	\end{subfigure}
	\caption{Varying $N_g$ (Number of gold tasks), Settings : Uniform Costs, $\beta = 0.01, \alpha = 0.01$}
	\label{fig:results-1}
\end{figure*}

\begin{figure*}
	\centering
	\begin{subfigure}{0.33\textwidth}
		\centering
		\includegraphics[width=1\textwidth, trim={20 20 14 20},clip]{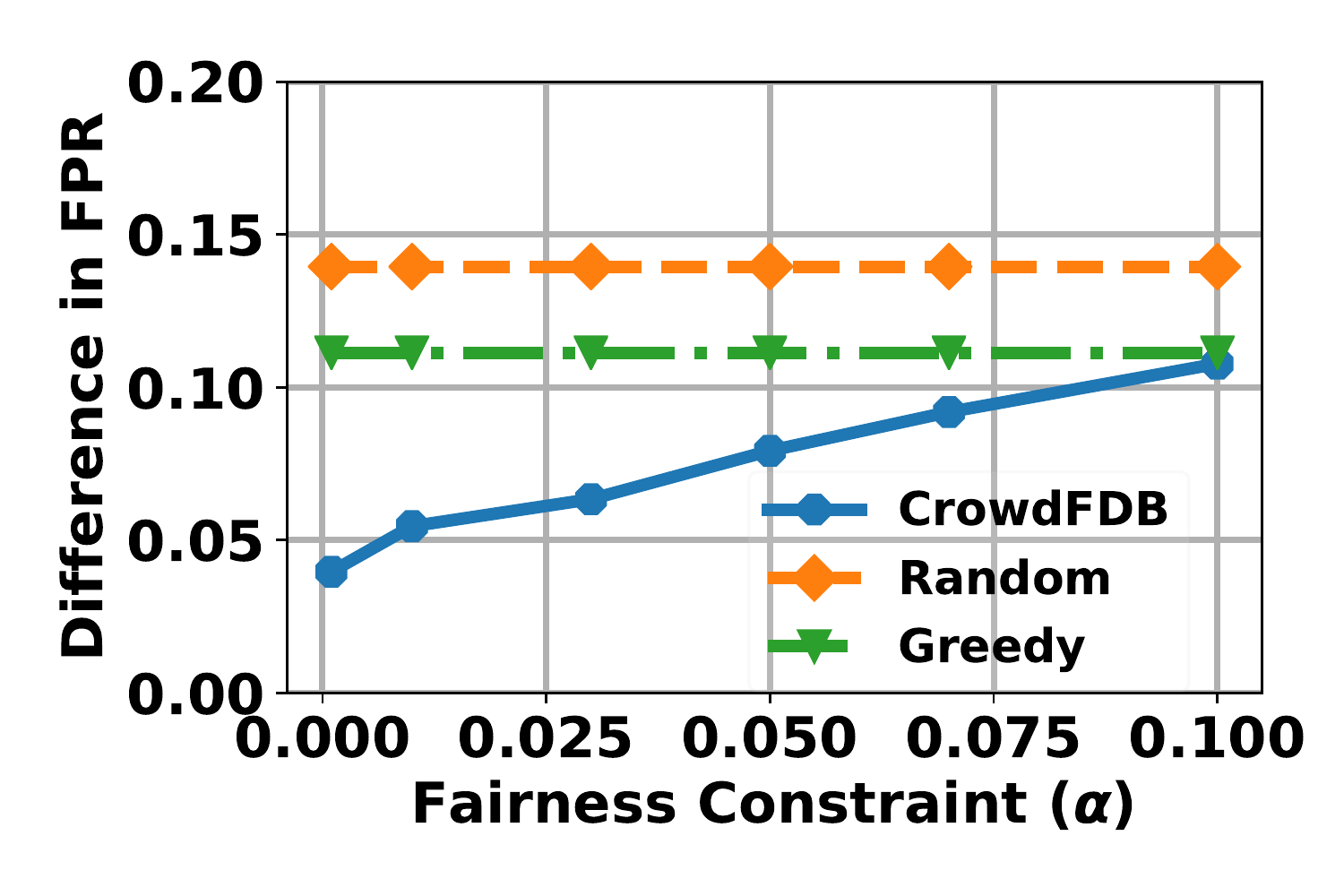}
		\caption{Absolute Difference in FPR}
		\label{fig:fpr-2} 
	\end{subfigure}
	\begin{subfigure}{0.33\textwidth}
		\centering
		\includegraphics[width=1\textwidth, trim={20 20 14 20},clip]{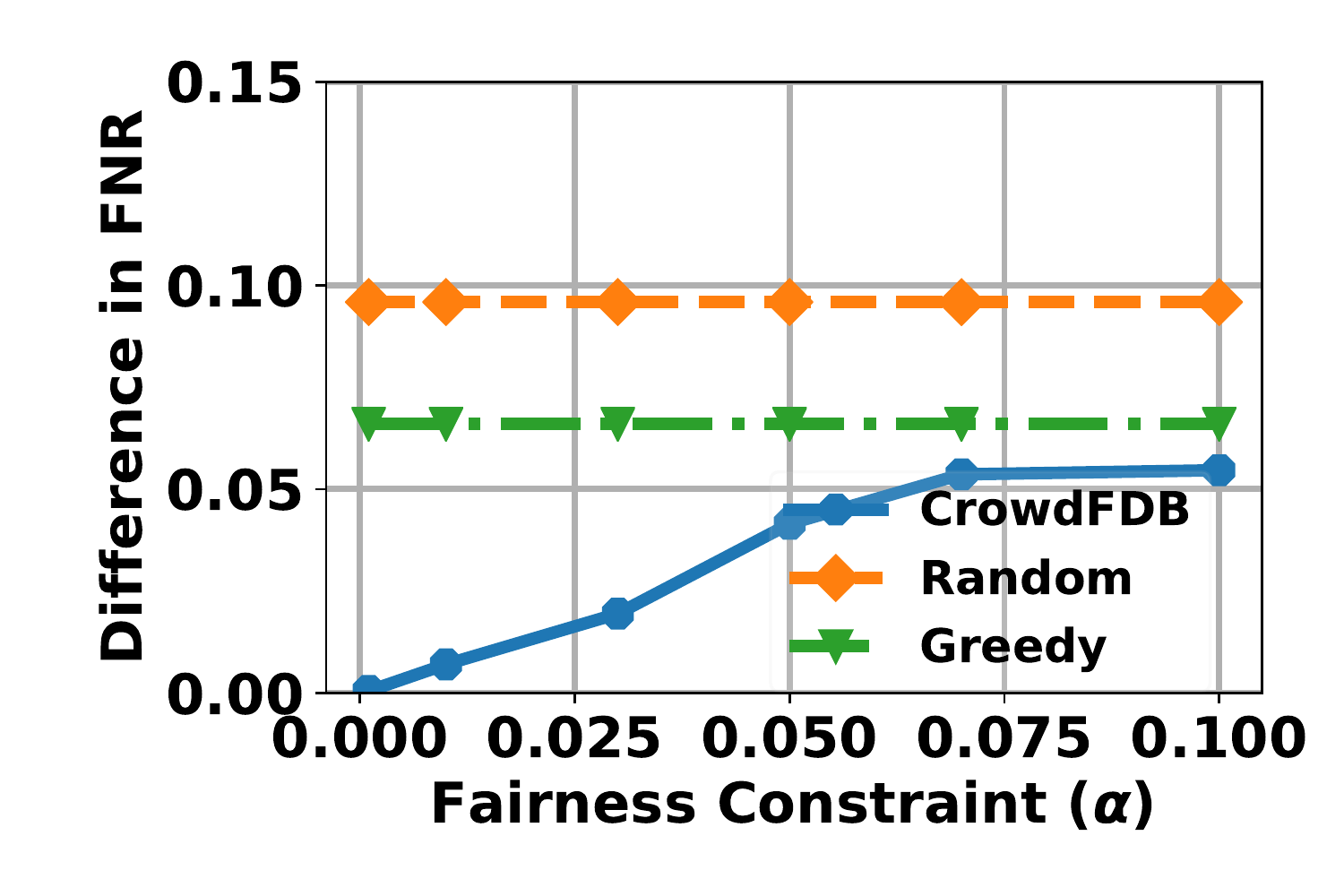}
		\caption{Absolute Difference in FNR}
		\label{fig:fnr-2} 
	\end{subfigure}
	\begin{subfigure}{0.33\textwidth}
		\centering
		\includegraphics[width=1\textwidth, trim={20 20 14 20},clip]{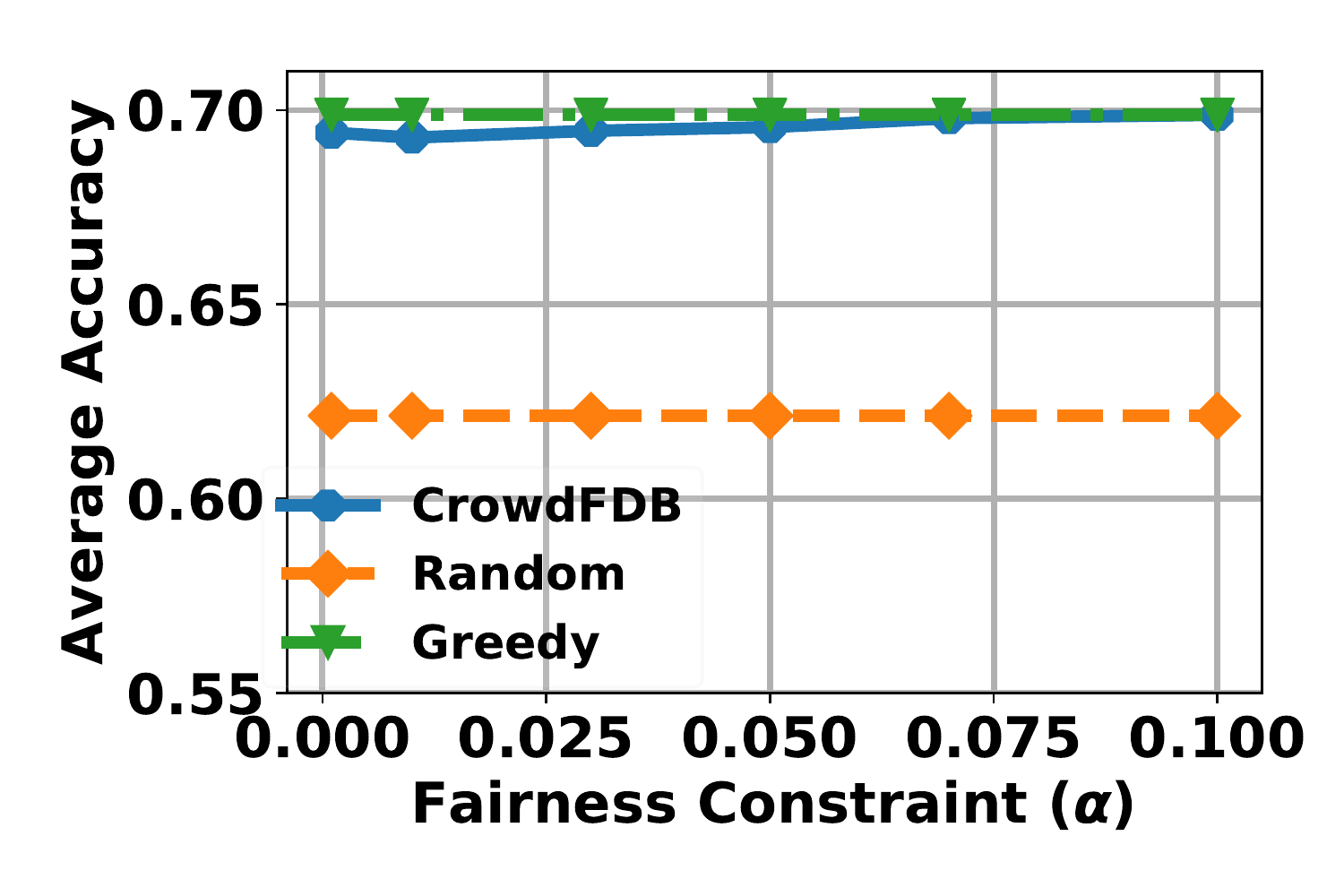}
		\caption{Average Accuracy}
		\label{fig:acc-2} 
	\end{subfigure}
	\caption{Varying $\alpha$ (Fairness Constraint), Settings : Uniform Costs, $\beta = 0.01, N_g = 20$}
	\label{fig:results-2}
\end{figure*}
\section{Theoretical Analysis}
When worker accuracy matrices are known, our method is guaranteed to provide the optimal solution, satisfying constraints. However, when estimates of the accuracy matrices are used, two interesting questions arise:
\begin{enumerate}
	\item Does the solution of problem~\ref{eq:optimization-3} (which is optimal and satisfies fairness constraints only according to the estimated accuracy parameters) also satisfy fairness in reality?
	\item How much does the requester lose in terms of actual expected accuracy of the policy because of using the estimated accuracy parameters in optimization?
\end{enumerate}
\begin{theorem}\label{thm:delta}
	With probability at least $\gamma$, the solution  $\hat{S}$ to the optimization problem~\ref{eq:optimization-3} satisfies
	\begin{equation*}
		\begin{split}
			\abs{\mathcal{A}_0[0,1] - \mathcal{A}_1[0,1]} \leq \alpha + \delta
		\end{split}
	\end{equation*}
	where \begin{small}$$\delta = 2\sqrt{\frac{ -\ln{(1-\sqrt[2n]\gamma)} + \ln{2}}{2N_{g}}} \text{;} \mathcal{A}_z = \sum_{i = 1}^{n} \hat{S}[i] \mathcal{A}_{iz},\forall z \in \{0,1\}$$\end{small} and $N_g$ is number of gold tasks.
\end{theorem}

The theorem states that when we use estimates of the worker accuracy matrices instead of the real matrices, the obtained solution $\hat{S}$ doesn't violate the fairness constraints in reality by more than $\delta$, with probability at least $\gamma$.

\begin{theorem}\label{thm:acc}
	Assuming that the optimal solution $\hat{S}$ of problem~\ref{eq:optimization-3} satisfies fairness constraints of problem~\ref{eq:optimization-2} and the optimal solution $S$ of problem~\ref{eq:optimization-2} satisfies fairness constraints of problem~\ref{eq:optimization-3}, then with probability at least $\gamma^\prime$
	$$\mathbbm{E}[\mathcal{A}(S)] - \mathbbm{E}[\mathcal{A}(\hat{S})] \leq 2n\beta\sqrt{\frac{-\ln{(1-\sqrt[4n]{\gamma^\prime})} + \ln{2}}{2N_{g}}}$$
	
	where \begin{small}$$\mathbbm{E}[\mathcal{A}(S)] = \kern-0.5em\sum_{z \in \{0,1\}} \kern-0.5em P(Z= z)\kern-0.5em\sum_{y \in \{0,1\}} \kern-0.5emP_z(Y=y) \sum_{i = 1}^{n} S[i] \mathcal{A}_{iz}[y,y],$$
		$$\mathbbm{E}[\mathcal{A}(\hat{S})] = \kern-0.5em\sum_{z \in \{0,1\}} \kern-0.5em P(Z= z)\kern-0.5em\sum_{y \in \{0,1\}} \kern-0.5emP_z(Y=y) \sum_{i = 1}^{n} \hat{S}[i] \mathcal{A}_{iz}[y,y]$$\end{small}
\end{theorem}

The theorem provides an upper bound on the loss in real expected accuracy of the crowdsourcing policy, when we use the estimated worker matrices instead of the real accuracy matrices for optimization. Note that in both the theorems, the bounds get better with increasing number of gold tasks.

The proofs of the above theorem are not difficult and depend on a simple application of the Hoeffding's inequality.
\section{Experimental Evaluation}
\begin{figure*}
	\centering
	\begin{subfigure}{0.33\textwidth}
		\centering
		\includegraphics[width=1\textwidth, trim={20 20 14 20},clip]{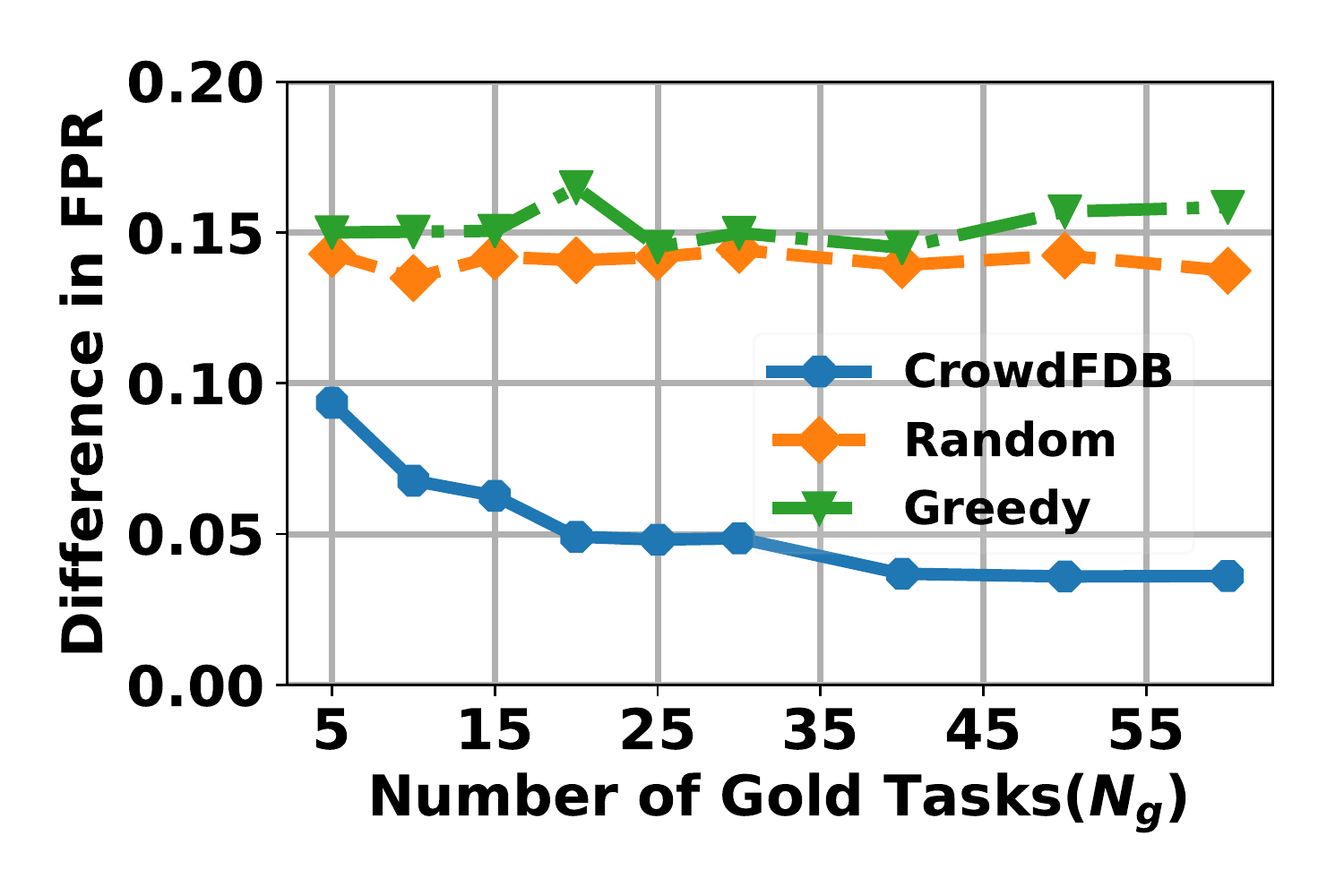}
		\caption{Absolute Difference in FPR}
		\label{fig:fpr-3} 
	\end{subfigure}
	\begin{subfigure}{0.33\textwidth}
		\centering
		\includegraphics[width=1\textwidth, trim={20 20 14 20},clip]{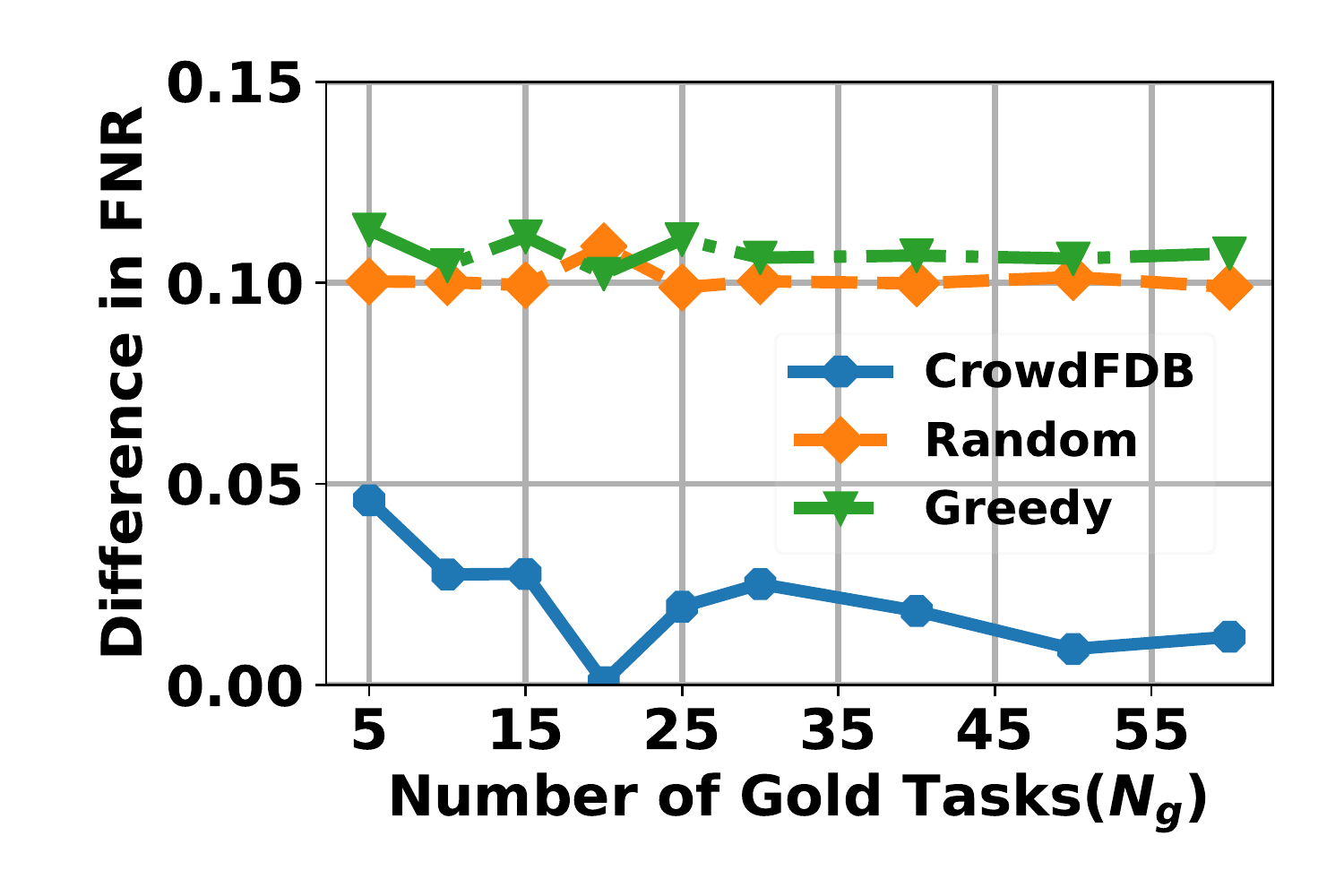}
		\caption{Absolute Difference in FNR}
		\label{fig:fnr-3} 
	\end{subfigure}
	\begin{subfigure}{0.33\textwidth}
		\centering
		\includegraphics[width=1\textwidth, trim={20 20 14 20},clip]{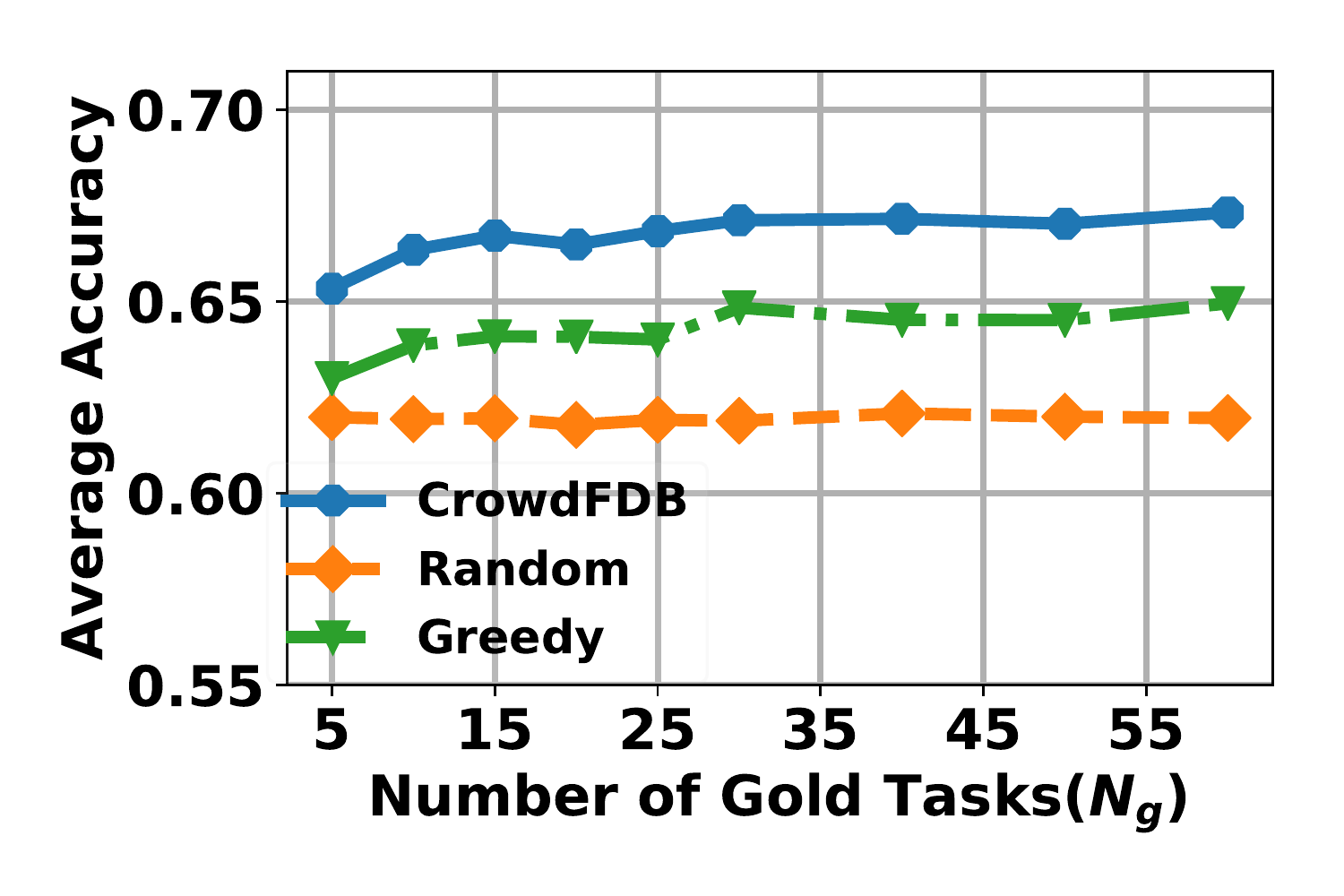}
		\caption{Average Accuracy}
		\label{fig:acc-3} 
	\end{subfigure}
	\caption{Varying $N_g$ (Number of gold tasks), Settings : Non-Uniform Costs, $\beta = 0.01, \alpha = 0.01, C = 1.5$}
	\label{fig:results-3}
	\centering
	\begin{subfigure}{0.33\textwidth}
		\centering
		\includegraphics[width=1\textwidth, trim={20 20 14 20},clip]{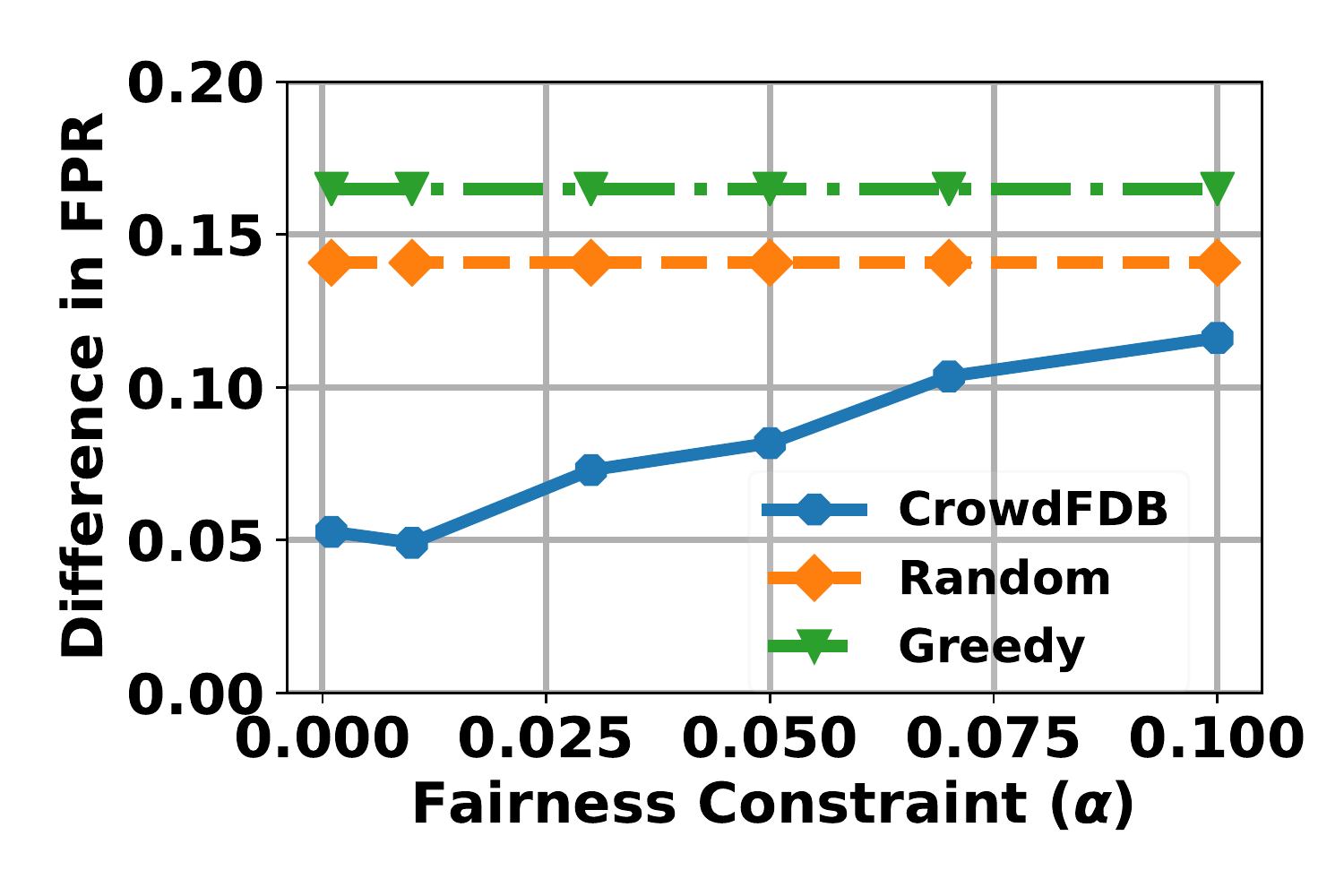}
		\caption{Absolute Difference in FPR}
		\label{fig:fpr-4} 
	\end{subfigure}
	\begin{subfigure}{0.33\textwidth}
		\centering
		\includegraphics[width=1\textwidth, trim={20 20 14 20},clip]{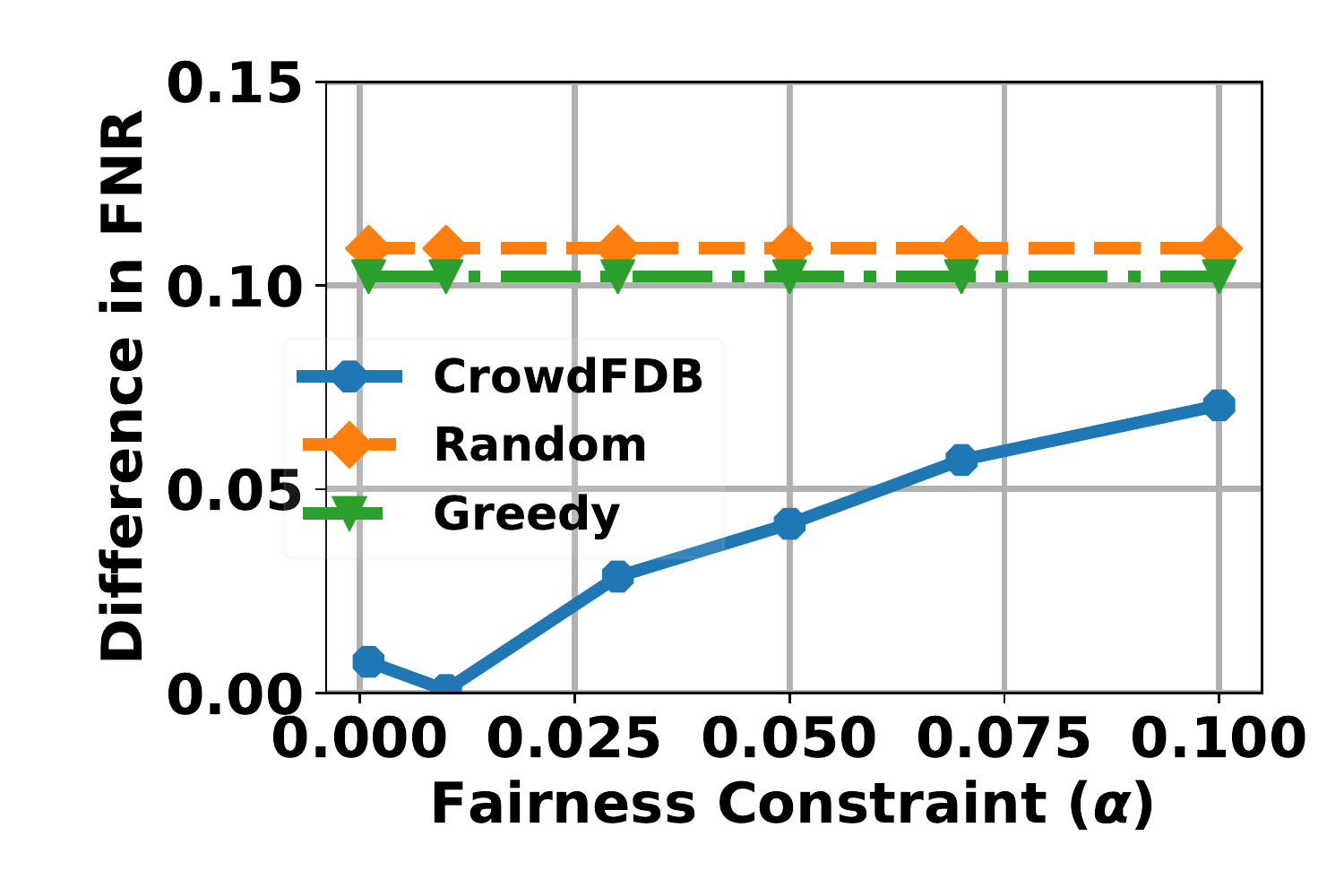}
		\caption{Absolute Difference in FNR}
		\label{fig:fnr-4} 
	\end{subfigure}
	\begin{subfigure}{0.33\textwidth}
		\centering
		\includegraphics[width=1\textwidth, trim={20 20 14 20},clip]{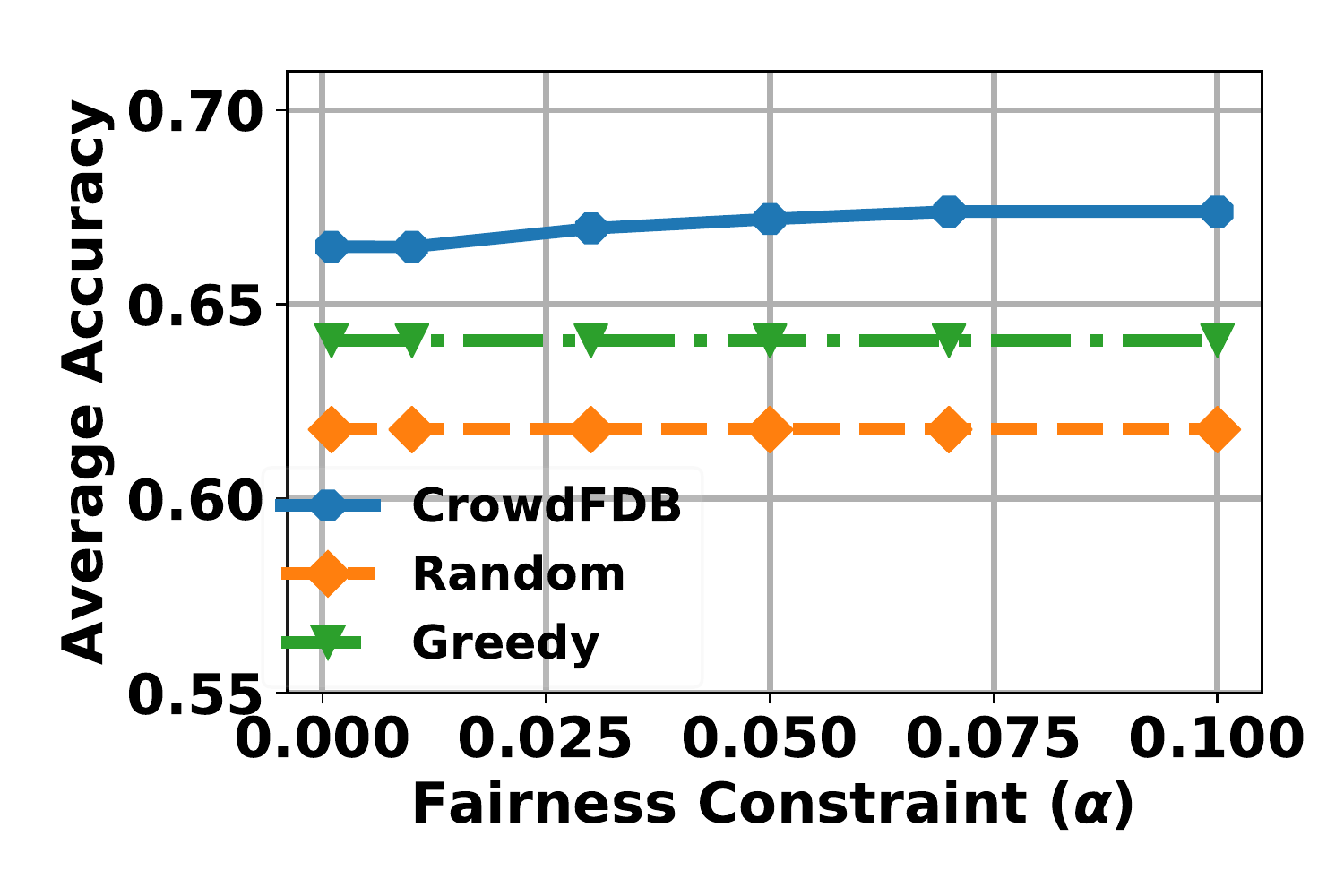}
		\caption{Average Accuracy}
		\label{fig:acc-4} 
	\end{subfigure}
	\caption{Varying $\alpha$ (Fairness Constraint), Settings : Non-Uniform Costs, $\beta = 0.01, N_g = 20, C = 1.5$}
	\label{fig:results-4}
\end{figure*}
\subsubsection{Datasets}\hspace*{-0.1cm}We use the following datasets in our experiments.
\begin{enumerate}
	\item \textbf{Broward County Dataset}~\cite{pro-publica} : This dataset contains information about 7214 defendants arrested in Broward County, Florida between 2013 and 2014. The information includes race of the defendants among other non-sensitive attributes such as age, prior charges etc. The dataset also contains ground-truth whether the defendants recidivated within 2 years or not. There are 3696 black defendants and 2454 white defendants in the dataset and the base rate of recidivism is 51.43\% among the black defendants and 39.36\% among the white defendants.
	
	\item \textbf{Crowd Judgment Dataset} : \cite{dressel2018accuracy} randomly selected a subset of 1000  defendants from the Broward County dataset and asked 20 random workers on Amazon Mechanical Turk to predict recidivism for each individual. In total, 400 workers participated in their study and each worker submitted answers for 50 different defendants. The dataset contains these answers collected from the crowd.
\end{enumerate}

\subsubsection{Experiment Outline} The idea is to split the set of defendants into two sets. The first set acts as the gold standard set, which we use to estimate worker accuracy matrices. Once we have the estimates of the worker accuracy matrices, we can solve the optimization problem~\ref{eq:optimization-3} and learn optimal crowdsourcing policy. We then pick non-gold defendants one by one and assign it to one of the 400 workers, randomly selected according to the policy. The workers' responses are then compared with the ground-truth label to evaluate fairness and accuracy of our crowdsourcing policy.

\subsubsection{Handling Limitations of Datasets} Unfortunately, none of the two datasets alone can be used for such experiment. The Broward County dataset contains ground truth labels but doesn't contain workers' answers. On the other hand, the Crowd Judgment dataset does contain worker answers but is very limited for the following reasons. In this dataset, tasks have already been assigned (randomly) to workers and for every defendant, we have responses of only a subset of 20 workers out of all 400 workers. If the crowdsourcing policy learned by our algorithm decides that a worker outside that subset of 20 workers should be assigned a task, then we will need to know the answer of that worker but the answer of this worker is not part of the dataset. The second reason is that every worker has submitted answers for 50 defendants, which is sufficient for getting good estimates of the accuracy parameters of the workers but not big enough to be further split into gold and non-gold sets.

\noindent To overcome these limitations, we first create a bigger synthetic dataset using the two real datasets as follows. We generate synthetic answers of all the 400 workers for all the 3696 black and 2454 white defendants in the Broward County dataset. The answers are generated using the worker accuracy parameters estimated from the entire Crowd Judgment dataset. Note that even though this is a synthetic dataset but none of the parameters of the dataset are synthetic. The worker accuracy parameters are derived from the entire real dataset of~\cite{dressel2018accuracy} and the base dataset (Broward County dataset) is used as it is. Indeed, this is not ideal but is perhaps the only option, given the limitations of the available datasets.

\noindent \underline{Worker Costs} : The datasets also don't contain workers' costs. We create this information in two different ways. In the first setting, we associate a uniform cost of $\$1$ to each worker. In the second and more interesting setting, we probabilistically associate a cost of $\$1$ or $\$3$. The probability of a worker's cost being $\$3$ is equal to her average accuracy and of it being $\$1$ is equal to $1-$her average accuracy. Thus, the higher the average accuracy of a worker, the higher is the probability that she will charge a cost of $\$3$.

Now this complete dataset is ready to be used in the experiment outlined earlier in this section. We compare our approach (called `CrowdFDB' in the figures) with two baselines (called `Random' and `Greedy'~\cite{tran2014efficient}). The baselines are discussed in appendix C.
\subsection{Observations} We use equal error rate parity (Definition~\ref{def:eqerror}) as the desired fairness. All results reported in the paper are averages over 100 repeated runs. Parameter $\beta$ was set to $0.01$ in the first set of experiments.  In the uniform costs settings, $C$ was set to $\$1$ and in non-uniform settings, $C = \$1.5$. 
\vspace{-0.2cm}
\subsubsection{Uniform Costs} In Figure~\ref{fig:results-1}, we keep the fairness constraint $\alpha$ to be fixed (0.01) and observe the effect of increasing number of gold tasks ($N_g$). Figures~\ref{fig:fpr-1} and~\ref{fig:fnr-1} show that as we increase $N_g$, the fairness i.e. the absolute difference in FPR (and FNR) for black and white populations, gets closer and closer to $\alpha$. In other words, the $\delta$ of Theorem~\ref{thm:delta} gets closer to $0$ as expected. Moreover, the margin between our algorithm and the baselines also increases. However, meeting the fairness constraints alone is not enough. This could also be done by a bad algorithm that collects equally wrong labels for both white and black populations. Hence, accuracy of the collected labels is also an important measure. Figure~\ref{fig:acc-1} shows that our algorithm has an accuracy competitive to the Greedy baseline method, which is a highly efficient baseline in the literature for accuracy optimization. Our algorithm can achieve same level of accuracy while also providing fairness. In Figure~\ref{fig:results-2}, we keep $N_g$ fixed (20) and observe the effect of increasing value of $\alpha$. As value of $\alpha$ increases, the fairness constraints are more relaxed and the algorithm can obtain better accuracy.
\vspace{-0.2cm}
\subsubsection{Non-Uniform Costs} In the non-uniform costs settings, we observe similar patterns in Figure~\ref{fig:results-3} and~\ref{fig:results-4}. There are a few notable differences. The accuracy of our algorithm as well as the Greedy baseline are lower. Our algorithm doesn't select more accurate workers because of budget constraints and the Greedy baseline also finds the density of the more accurate workers comparatively lower due to their high costs and prefers to choose other high density workers. In this case, our algorithm beats Greedy in not just fairness but also in accuracy by better utilizing the available budget.

Figures~\ref{fig:results-1_} and \ref{fig:results-2_} of appendix D show the results with $\beta = 0.01, C = 2.5$. Figures~\ref{fig:results-3_} and \ref{fig:results-4_} of appendix D show the results with $\beta = 0.005, C = 2.5$. Decreasing the value of parameter $\beta$ makes the algorithms (ours and the Greedy baseline) more constrained in assigning tasks to the workers that they find to be better. This hits the accuracy of both the algorithms but the general trends discussed above (w.r.t. fairness and accuracy with different $N_g$ and $\alpha$) remain the same. The effect of increasing budget from $1.5$ to $2.5$ is that our algorithm can get better accuracy but there is no effect on the performance of other baselines as expected, since there was no explicit budget constraint placed on them.
\vspace{-0.2cm}
\section{Conclusions and Future Work}
In this paper, we addressed the problem of data fairness in crowdsourcing. We proposed a novel crowdsourcing algorithm that learns an optimal selection probability distribution over the available set of workers to maximize the expected accuracy of collected data, while ensuring that the errors in the data are not unfairly discriminatory towards any particular social group. There also remain many challenges to be addressed. These include estimating accuracy without requiring gold standard tasks, relaxing the assumption about knowledge of prior label distribution and the i.i.d. assumption about workers' answers.

Another interesting challenge in slightly different settings is to define data fairness for the case of subjective tasks, which have no ground truth labels and thus, no clear notion of errors. Ensuring fairness in subjective data collection is also likely to create a challenging problem of lying incentives\cite{goel2019} for workers.
%
\bibliographystyle{ACM-Reference-Format}
\bibliography{sample-base}


\begin{thebibliography}{30}


\ifx \showCODEN    \undefined \def \showCODEN     #1{\unskip}     \fi
\ifx \showDOI      \undefined \def \showDOI       #1{#1}\fi
\ifx \showISBNx    \undefined \def \showISBNx     #1{\unskip}     \fi
\ifx \showISBNxiii \undefined \def \showISBNxiii  #1{\unskip}     \fi
\ifx \showISSN     \undefined \def \showISSN      #1{\unskip}     \fi
\ifx \showLCCN     \undefined \def \showLCCN      #1{\unskip}     \fi
\ifx \shownote     \undefined \def \shownote      #1{#1}          \fi
\ifx \showarticletitle \undefined \def \showarticletitle #1{#1}   \fi
\ifx \showURL      \undefined \def \showURL       {\relax}        \fi
\providecommand\bibfield[2]{#2}
\providecommand\bibinfo[2]{#2}
\providecommand\natexlab[1]{#1}
\providecommand\showeprint[2][]{arXiv:#2}

\bibitem[\protect\citeauthoryear{Barocas, Hardt, and Narayanan}{Barocas
  et~al\mbox{.}}{2018}]%
        {fairness-book}
\bibfield{author}{\bibinfo{person}{Solon Barocas}, \bibinfo{person}{Moritz
  Hardt}, {and} \bibinfo{person}{Arvind Narayanan}.}
  \bibinfo{year}{2018}\natexlab{}.
\newblock \showarticletitle{Fairness and machine learning : Limitations and
  Opportunities}.
\newblock \bibinfo{howpublished}{\url{http://fairmlbook.org}}.
\newblock  (\bibinfo{year}{2018}).
\newblock


\bibitem[\protect\citeauthoryear{Barocas and Selbst}{Barocas and
  Selbst}{2016}]%
        {barocas2016big}
\bibfield{author}{\bibinfo{person}{Solon Barocas} {and}
  \bibinfo{person}{Andrew~D Selbst}.} \bibinfo{year}{2016}\natexlab{}.
\newblock \showarticletitle{Big data's disparate impact}.
\newblock \bibinfo{journal}{\emph{California Law Review 671}}
  (\bibinfo{year}{2016}).
\newblock


\bibitem[\protect\citeauthoryear{Bolukbasi, Chang, Zou, Saligrama, and
  Kalai}{Bolukbasi et~al\mbox{.}}{2016}]%
        {bolukbasi2016man}
\bibfield{author}{\bibinfo{person}{Tolga Bolukbasi}, \bibinfo{person}{Kai-Wei
  Chang}, \bibinfo{person}{James~Y Zou}, \bibinfo{person}{Venkatesh Saligrama},
  {and} \bibinfo{person}{Adam~T Kalai}.} \bibinfo{year}{2016}\natexlab{}.
\newblock \showarticletitle{Man is to computer programmer as woman is to
  homemaker? debiasing word embeddings}. In \bibinfo{booktitle}{\emph{Advances
  in Neural Information Processing Systems}}.
\newblock


\bibitem[\protect\citeauthoryear{Bragg, Weld, et~al\mbox{.}}{Bragg
  et~al\mbox{.}}{2016}]%
        {bragg2016optimal}
\bibfield{author}{\bibinfo{person}{Jonathan Bragg}, \bibinfo{person}{Daniel~S
  Weld}, {et~al\mbox{.}}} \bibinfo{year}{2016}\natexlab{}.
\newblock \showarticletitle{Optimal testing for crowd workers}. In
  \bibinfo{booktitle}{\emph{International Conference on Autonomous Agents \&
  Multiagent Systems}}.
\newblock


\bibitem[\protect\citeauthoryear{Calmon, Wei, Vinzamuri, Ramamurthy, and
  Varshney}{Calmon et~al\mbox{.}}{2017}]%
        {calmon2017optimized}
\bibfield{author}{\bibinfo{person}{Flavio Calmon}, \bibinfo{person}{Dennis
  Wei}, \bibinfo{person}{Bhanukiran Vinzamuri},
  \bibinfo{person}{Karthikeyan~Natesan Ramamurthy}, {and}
  \bibinfo{person}{Kush~R Varshney}.} \bibinfo{year}{2017}\natexlab{}.
\newblock \showarticletitle{Optimized Pre-Processing for Discrimination
  Prevention}. In \bibinfo{booktitle}{\emph{Advances in Neural Information
  Processing Systems}}.
\newblock


\bibitem[\protect\citeauthoryear{Chvatal}{Chvatal}{1983}]%
        {chvatal1983linear}
\bibfield{author}{\bibinfo{person}{Vasek Chvatal}.}
  \bibinfo{year}{1983}\natexlab{}.
\newblock \bibinfo{booktitle}{\emph{Linear programming}}.
\newblock \bibinfo{publisher}{Macmillan}.
\newblock


\bibitem[\protect\citeauthoryear{Dawid and Skene}{Dawid and Skene}{1979}]%
        {dawid1979maximum}
\bibfield{author}{\bibinfo{person}{Alexander~Philip Dawid} {and}
  \bibinfo{person}{Allan~M Skene}.} \bibinfo{year}{1979}\natexlab{}.
\newblock \showarticletitle{Maximum likelihood estimation of observer
  error-rates using the EM algorithm}.
\newblock \bibinfo{journal}{\emph{Applied statistics}} (\bibinfo{year}{1979}).
\newblock


\bibitem[\protect\citeauthoryear{Dressel and Farid}{Dressel and Farid}{2018}]%
        {dressel2018accuracy}
\bibfield{author}{\bibinfo{person}{Julia Dressel} {and} \bibinfo{person}{Hany
  Farid}.} \bibinfo{year}{2018}\natexlab{}.
\newblock \showarticletitle{The accuracy, fairness, and limits of predicting
  recidivism}.
\newblock \bibinfo{journal}{\emph{Science advances}} (\bibinfo{year}{2018}).
\newblock


\bibitem[\protect\citeauthoryear{Dwork, Hardt, Pitassi, Reingold, and
  Zemel}{Dwork et~al\mbox{.}}{2012}]%
        {dwork2012fairness}
\bibfield{author}{\bibinfo{person}{Cynthia Dwork}, \bibinfo{person}{Moritz
  Hardt}, \bibinfo{person}{Toniann Pitassi}, \bibinfo{person}{Omer Reingold},
  {and} \bibinfo{person}{Richard Zemel}.} \bibinfo{year}{2012}\natexlab{}.
\newblock \showarticletitle{Fairness through awareness}. In
  \bibinfo{booktitle}{\emph{Proceedings of ITCS}}.
\newblock


\bibitem[\protect\citeauthoryear{Feldman, Friedler, Moeller, Scheidegger, and
  Venkatasubramanian}{Feldman et~al\mbox{.}}{2015}]%
        {feldman2015certifying}
\bibfield{author}{\bibinfo{person}{Michael Feldman}, \bibinfo{person}{Sorelle~A
  Friedler}, \bibinfo{person}{John Moeller}, \bibinfo{person}{Carlos
  Scheidegger}, {and} \bibinfo{person}{Suresh Venkatasubramanian}.}
  \bibinfo{year}{2015}\natexlab{}.
\newblock \showarticletitle{Certifying and removing disparate impact}. In
  \bibinfo{booktitle}{\emph{21th ACM SIGKDD International Conference on
  Knowledge Discovery and Data Mining}}.
\newblock


\bibitem[\protect\citeauthoryear{Goel and Faltings}{Goel and Faltings}{2019}]%
        {goel2019}
\bibfield{author}{\bibinfo{person}{Naman Goel} {and} \bibinfo{person}{Boi
  Faltings}.} \bibinfo{year}{2019}\natexlab{}.
\newblock \showarticletitle{Deep Bayesian Trust: A Dominant and Fair Incentive
  Mechanism for Crowd}. In \bibinfo{booktitle}{\emph{AAAI Conference on
  Artificial Intelligence}}.
\newblock


\bibitem[\protect\citeauthoryear{Goel, Yaghini, and Faltings}{Goel
  et~al\mbox{.}}{2018}]%
        {goel2018non}
\bibfield{author}{\bibinfo{person}{Naman Goel}, \bibinfo{person}{Mohammad
  Yaghini}, {and} \bibinfo{person}{Boi Faltings}.}
  \bibinfo{year}{2018}\natexlab{}.
\newblock \showarticletitle{Non-discriminatory machine learning through convex
  fairness criteria}. In \bibinfo{booktitle}{\emph{Thirty-Second AAAI
  Conference on Artificial Intelligence}}.
\newblock


\bibitem[\protect\citeauthoryear{Hann{\'a}k, Wagner, Garcia, Mislove,
  Strohmaier, and Wilson}{Hann{\'a}k et~al\mbox{.}}{2017}]%
        {hannak2017bias}
\bibfield{author}{\bibinfo{person}{Anik{\'o} Hann{\'a}k},
  \bibinfo{person}{Claudia Wagner}, \bibinfo{person}{David Garcia},
  \bibinfo{person}{Alan Mislove}, \bibinfo{person}{Markus Strohmaier}, {and}
  \bibinfo{person}{Christo Wilson}.} \bibinfo{year}{2017}\natexlab{}.
\newblock \showarticletitle{Bias in Online Freelance Marketplaces: Evidence
  from TaskRabbit and Fiverr.}. In \bibinfo{booktitle}{\emph{CSCW}}.
  \bibinfo{pages}{1914--1933}.
\newblock


\bibitem[\protect\citeauthoryear{Hardt, Price, Srebro, et~al\mbox{.}}{Hardt
  et~al\mbox{.}}{2016}]%
        {hardt2016equality}
\bibfield{author}{\bibinfo{person}{Moritz Hardt}, \bibinfo{person}{Eric Price},
  \bibinfo{person}{Nati Srebro}, {et~al\mbox{.}}}
  \bibinfo{year}{2016}\natexlab{}.
\newblock \showarticletitle{Equality of opportunity in supervised learning}. In
  \bibinfo{booktitle}{\emph{Advances in Neural Information Processing
  Systems}}.
\newblock


\bibitem[\protect\citeauthoryear{Ho, Jabbari, and Vaughan}{Ho
  et~al\mbox{.}}{2013}]%
        {ho2013adaptive}
\bibfield{author}{\bibinfo{person}{Chien-Ju Ho}, \bibinfo{person}{Shahin
  Jabbari}, {and} \bibinfo{person}{Jennifer~Wortman Vaughan}.}
  \bibinfo{year}{2013}\natexlab{}.
\newblock \showarticletitle{Adaptive task assignment for crowdsourced
  classification}. In \bibinfo{booktitle}{\emph{Proceedings of ICML}}.
\newblock


\bibitem[\protect\citeauthoryear{Ho and Vaughan}{Ho and Vaughan}{2012}]%
        {ho2012online}
\bibfield{author}{\bibinfo{person}{Chien-Ju Ho} {and}
  \bibinfo{person}{Jennifer~Wortman Vaughan}.} \bibinfo{year}{2012}\natexlab{}.
\newblock \showarticletitle{Online task assignment in crowdsourcing markets}.
  In \bibinfo{booktitle}{\emph{Proceedings of AAAI Conference}}.
\newblock


\bibitem[\protect\citeauthoryear{Karger, Oh, and Shah}{Karger
  et~al\mbox{.}}{2014}]%
        {Karger:2014:BTA:2765242.2765244}
\bibfield{author}{\bibinfo{person}{David~R. Karger}, \bibinfo{person}{Sewoong
  Oh}, {and} \bibinfo{person}{Devavrat Shah}.} \bibinfo{year}{2014}\natexlab{}.
\newblock \showarticletitle{Budget-Optimal Task Allocation for Reliable
  Crowdsourcing Systems}.
\newblock \bibinfo{journal}{\emph{Opeations Research}} (\bibinfo{year}{2014}),
  \bibinfo{pages}{1--24}.
\newblock


\bibitem[\protect\citeauthoryear{Kay, Matuszek, and Munson}{Kay
  et~al\mbox{.}}{2015}]%
        {kay2015unequal}
\bibfield{author}{\bibinfo{person}{Matthew Kay}, \bibinfo{person}{Cynthia
  Matuszek}, {and} \bibinfo{person}{Sean~A Munson}.}
  \bibinfo{year}{2015}\natexlab{}.
\newblock \showarticletitle{Unequal representation and gender stereotypes in
  image search results for occupations}. In \bibinfo{booktitle}{\emph{33rd
  Annual ACM Conference on Human Factors in Computing Systems}}.
\newblock


\bibitem[\protect\citeauthoryear{Kleinberg, Mullainathan, and
  Raghavan}{Kleinberg et~al\mbox{.}}{2017}]%
        {Kleinberg2017InherentTI}
\bibfield{author}{\bibinfo{person}{Jon~M. Kleinberg}, \bibinfo{person}{Sendhil
  Mullainathan}, {and} \bibinfo{person}{Manish Raghavan}.}
  \bibinfo{year}{2017}\natexlab{}.
\newblock \showarticletitle{Inherent Trade-Offs in the Fair Determination of
  Risk Scores}. In \bibinfo{booktitle}{\emph{Proceedings of ITCS}}.
\newblock


\bibitem[\protect\citeauthoryear{Kulshrestha, Eslami, Messias, Zafar, Ghosh,
  Gummadi, and Karahalios}{Kulshrestha et~al\mbox{.}}{2017}]%
        {kulshrestha2017quantifying}
\bibfield{author}{\bibinfo{person}{Juhi Kulshrestha},
  \bibinfo{person}{Motahhare Eslami}, \bibinfo{person}{Johnnatan Messias},
  \bibinfo{person}{Muhammad~Bilal Zafar}, \bibinfo{person}{Saptarshi Ghosh},
  \bibinfo{person}{Krishna~P Gummadi}, {and} \bibinfo{person}{Karrie
  Karahalios}.} \bibinfo{year}{2017}\natexlab{}.
\newblock \showarticletitle{Quantifying search bias: Investigating sources of
  bias for political searches in social media}. In
  \bibinfo{booktitle}{\emph{ACM Conference on Computer Supported Cooperative
  Work and Social Computing}}.
\newblock


\bibitem[\protect\citeauthoryear{Neel and Roth}{Neel and Roth}{2018}]%
        {sethneel}
\bibfield{author}{\bibinfo{person}{Seth Neel} {and} \bibinfo{person}{Aaron
  Roth}.} \bibinfo{year}{2018}\natexlab{}.
\newblock \showarticletitle{Mitigating Bias in Adaptive Data Gathering via
  Differential Privacy}. In \bibinfo{booktitle}{\emph{35th International
  Conference on Machine Learning}}.
\newblock


\bibitem[\protect\citeauthoryear{Oleson, Sorokin, Laughlin, Hester, Le, and
  Biewald}{Oleson et~al\mbox{.}}{2011}]%
        {oleson2011}
\bibfield{author}{\bibinfo{person}{David Oleson}, \bibinfo{person}{Alexander
  Sorokin}, \bibinfo{person}{Greg Laughlin}, \bibinfo{person}{Vaughn Hester},
  \bibinfo{person}{John Le}, {and} \bibinfo{person}{Lukas Biewald}.}
  \bibinfo{year}{2011}\natexlab{}.
\newblock \showarticletitle{Programmatic Gold: Targeted and Scalable Quality
  Assurance in Crowdsourcing}. In \bibinfo{booktitle}{\emph{AAAI Workshop on
  Human Computation}}.
\newblock


\bibitem[\protect\citeauthoryear{Otterbacher}{Otterbacher}{2015a}]%
        {otterbacher2015crowdsourcing}
\bibfield{author}{\bibinfo{person}{Jahna Otterbacher}.}
  \bibinfo{year}{2015}\natexlab{a}.
\newblock \showarticletitle{Crowdsourcing stereotypes: Linguistic bias in
  metadata generated via gwap}. In \bibinfo{booktitle}{\emph{Proceedings of ACM
  CHI}}.
\newblock


\bibitem[\protect\citeauthoryear{Otterbacher}{Otterbacher}{2015b}]%
        {otterbacher2015linguistic}
\bibfield{author}{\bibinfo{person}{Jahna Otterbacher}.}
  \bibinfo{year}{2015}\natexlab{b}.
\newblock \showarticletitle{Linguistic Bias in Collaboratively Produced
  Biographies: Crowdsourcing Social Stereotypes?}. In
  \bibinfo{booktitle}{\emph{Proceedings of AAAI ICWSM}}.
\newblock


\bibitem[\protect\citeauthoryear{Otterbacher, Bates, and Clough}{Otterbacher
  et~al\mbox{.}}{2017}]%
        {otterbacher2017competent}
\bibfield{author}{\bibinfo{person}{Jahna Otterbacher}, \bibinfo{person}{Jo
  Bates}, {and} \bibinfo{person}{Paul Clough}.}
  \bibinfo{year}{2017}\natexlab{}.
\newblock \showarticletitle{Competent men and warm women: Gender stereotypes
  and backlash in image search results}. In \bibinfo{booktitle}{\emph{ACM
  CHI}}.
\newblock


\bibitem[\protect\citeauthoryear{ProPublica}{ProPublica}{2017}]%
        {pro-publica}
\bibfield{author}{\bibinfo{person}{ProPublica}.}
  \bibinfo{year}{2017}\natexlab{}.
\newblock \showarticletitle{https://www.propublica.org/datastore/
  dataset/compas-recidivism-risk-score-data-and-analysis}.
\newblock  (\bibinfo{year}{2017}).
\newblock


\bibitem[\protect\citeauthoryear{Tran-Thanh, Stein, Rogers, and
  Jennings}{Tran-Thanh et~al\mbox{.}}{2014}]%
        {tran2014efficient}
\bibfield{author}{\bibinfo{person}{Long Tran-Thanh}, \bibinfo{person}{Sebastian
  Stein}, \bibinfo{person}{Alex Rogers}, {and} \bibinfo{person}{Nicholas~R
  Jennings}.} \bibinfo{year}{2014}\natexlab{}.
\newblock \showarticletitle{Efficient crowdsourcing of unknown experts using
  bounded multi-armed bandits}.
\newblock \bibinfo{journal}{\emph{Artificial Intelligence}}
  (\bibinfo{year}{2014}).
\newblock


\bibitem[\protect\citeauthoryear{Valera, Singla, and Rodriguez}{Valera
  et~al\mbox{.}}{2018}]%
        {valera2018enhancing}
\bibfield{author}{\bibinfo{person}{Isabel Valera}, \bibinfo{person}{Adish
  Singla}, {and} \bibinfo{person}{Manuel~Gomez Rodriguez}.}
  \bibinfo{year}{2018}\natexlab{}.
\newblock \showarticletitle{Enhancing the accuracy and fairness of human
  decision making}. In \bibinfo{booktitle}{\emph{Advances in Neural Information
  Processing Systems}}. \bibinfo{pages}{1774--1783}.
\newblock


\bibitem[\protect\citeauthoryear{{{White House}}}{{{White House}}}{2016}]%
        {president-statement}
\bibfield{author}{\bibinfo{person}{{{White House}}}.}
  \bibinfo{year}{2016}\natexlab{}.
\newblock \showarticletitle{Big Data: A Report on Algorithmic Systems,
  Opportunity, and Civil Rights.}
\newblock \bibinfo{journal}{\emph{Executive Office of the President}}
  (\bibinfo{year}{2016}).
\newblock


\bibitem[\protect\citeauthoryear{Zafar, Valera, Rodriguez, and Gummadi}{Zafar
  et~al\mbox{.}}{2017}]%
        {zafar2017fairness-www}
\bibfield{author}{\bibinfo{person}{Muhammad~Bilal Zafar},
  \bibinfo{person}{Isabel Valera}, \bibinfo{person}{Manuel~Gomez Rodriguez},
  {and} \bibinfo{person}{Krishna~P Gummadi}.} \bibinfo{year}{2017}\natexlab{}.
\newblock \showarticletitle{Fairness Constraints: Mechanisms for Fair
  Classification}. In \bibinfo{booktitle}{\emph{20th International Conference
  on Artificial Intelligence and Statistics (AISTATS)}}.
\newblock


\end{thebibliography}

%
\appendix
\vspace{-0.35cm}
\section{Estimated Accuracy Matrices}
Let $\hat{\mathcal{A}}_{iz}$ be the estimate of the worker accuracy matrices $\mathcal{A}_{iz},\  \forall z\in\{0,1\}$. If a worker $i$ answers $k$ tasks correctly out of $N_g$ gold tasks of type $z=1, y = 1$, then $$\hat{\mathcal{A}}_{i1}[1,1] = \frac{k}{N_g} \text{ and } \hat{\mathcal{A}}_{i1}[1,0] = 1-\hat{\mathcal{A}}_{i1}[1,1]$$
Similarly, if she answers $k^\prime$ tasks correctly out of $N_g$ gold tasks of type $z=1, y = 0$, then $$\hat{\mathcal{A}}_{i1}[0,0] = \frac{k^\prime}{N_g}\text{ and }\hat{\mathcal{A}}_{i1}[0,1] =1- \hat{\mathcal{A}}_{i1}[0,0]$$
This process is repeated with gold tasks of type $z = 0, y =1$ and $z = 0, y = 0$ to get estimates of all entries of the worker's matrices.
\vspace{-0.2cm}
\section{Summary of steps of the algorithm}
\vspace{-0.4cm}
\begin{algorithm}
	Assign $N_g$ gold tasks of every type ($(z = 0, y = 0), (z = 0, y = 1), (z = 1, y = 0), (z = 1, y = 1)$) to every worker $i$ in the provided set of $n$ workers. \\
	
	Get estimate of every worker $i$'s accuracy matrices $\hat{\mathcal{A}}_{i0}$ \& $\hat{\mathcal{A}}_{i1}$.\\
	
	Solve the linear program to find the best crowdsourcing policy satisfying desired fairness, diversity and budget constraints.\\
	
	Pick a task randomly from the task pool of tasks with unknown ground truth labels. Randomly select one or more workers from the set of $n$ workers, with probabilities specified by the crowdsourcing policy. Assign the task to the selected worker(s).\\
	
	Repeat Step 4 for all tasks in the pool.
	\caption{Crowdsourcing with Fairness, Diversity and Budget Constraints (CrowdFDB)}
	\label{alg:algo}
\end{algorithm}
\vspace{-0.8cm}
\section{Baselines} 
\begin{enumerate}
	\item \textbf{Random Policy} : In the random policy, all workers are equally likely to be selected (probability $\frac{1}{400}$) for any task.
	
	\item \textbf{Greedy Optimization~\cite{tran2014efficient}} : In this baseline, we first estimate worker accuracy matrices using gold tasks in exactly same way as we do in our algorithm. However, the optimization is done using a bounded knapsack algorithm instead of linear programming. This algorithm sorts the workers in decreasing order of their ``density", where density is defined as the ratio of the expected accuracy of a worker and her cost. The expected accuracy of a worker can be calculated in the same way as we do for our algorithm i.e. \begin{small}$\sum_{z \in \{0,1\}} P(Z= z)\sum_{g \in \{0,1\}} P_z(Y=y) \hat{\mathcal{A}}_{iz}[y,y]$\end{small}. Then, the algorithm assigns as many tasks as possible to the highest density worker without violating the diversity constraint. If $T$ is the total number of tasks to be assigned, then a worker can be assigned at most $\beta T$ tasks. Note that, unlike our algorithm, this baseline has to know the total number of tasks in advance to enforce diversity constraint. Once this worker has reached its capacity, the algorithm starts assigning tasks to the worker with next highest density and continues this for all tasks. However, it may be noted that the Greedy approach was originally proposed for a bit different setting, in which there is an overall crowdsourcing budget and the goal of the requester is to get as many tasks done as possible in that budget, maximizing total utility/accuracy and respecting work limits of the workers.
	
	To give extra advantage to the baselines, we don't put an \textit{explicit} budget constraint for them and observe whether they can compete with the fairness and accuracy of our algorithm, which operates under budget constraint.
\end{enumerate}

\section{Additional Experimental Results}
\begin{figure*}
	\centering
	\begin{subfigure}{0.33\textwidth}
		\centering
		\includegraphics[width=1\textwidth]{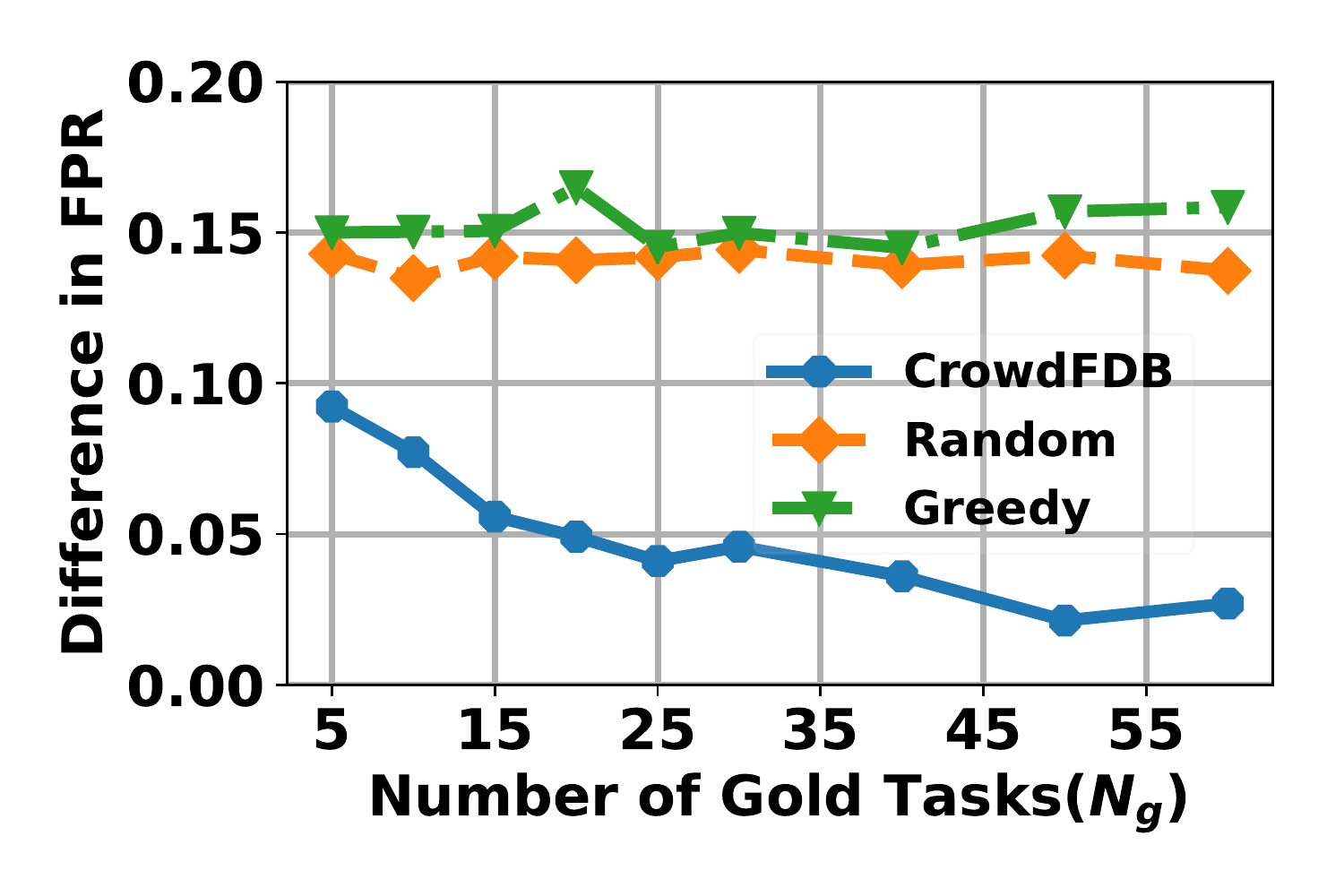}
		\caption{Absolute Difference in FPR}
		\label{fig:fpr-3_} 
	\end{subfigure}
	\begin{subfigure}{0.33\textwidth}
		\centering
		\includegraphics[width=1\textwidth]{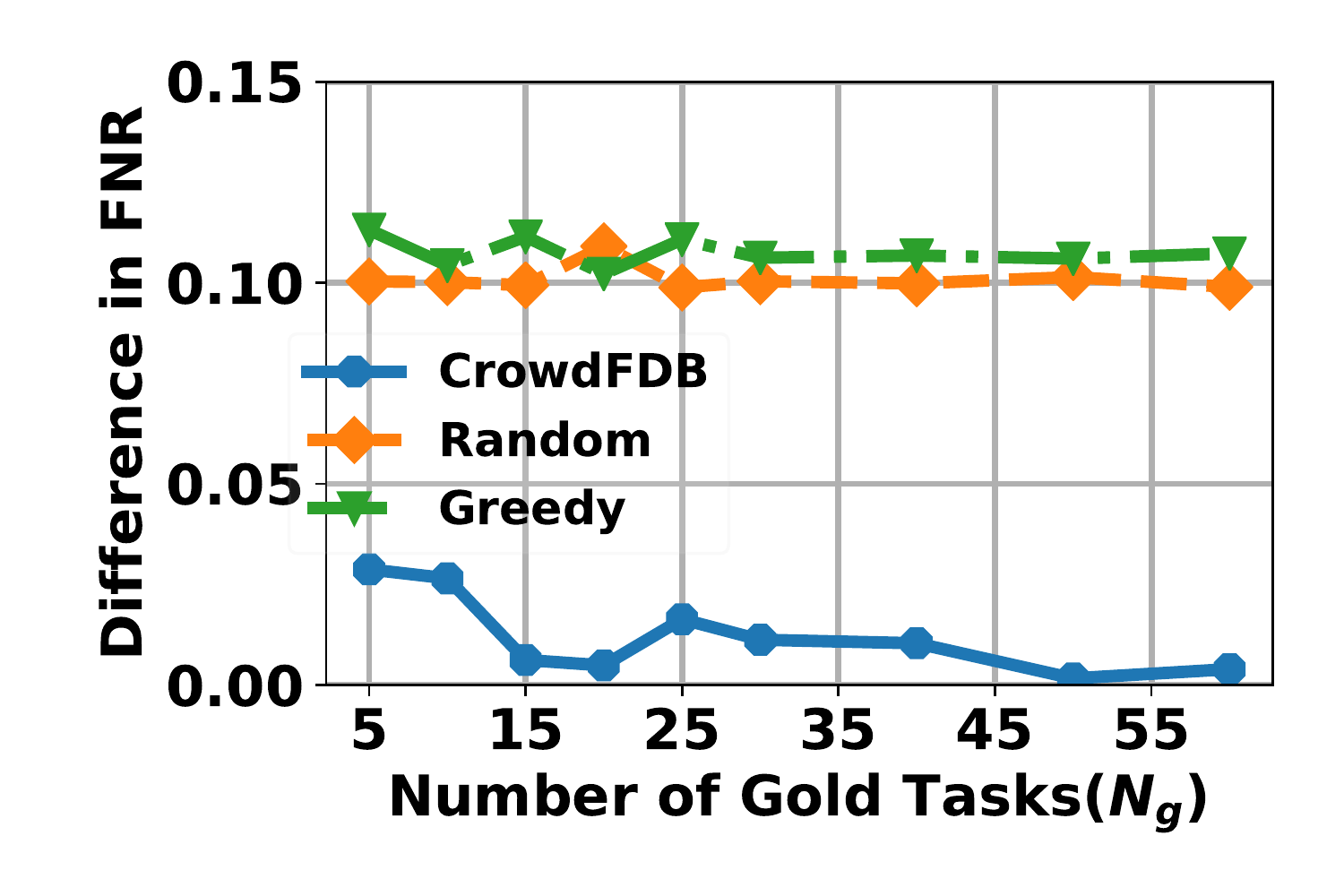}
		\caption{Absolute Difference in FNR}
		\label{fig:fnr-3_} 
	\end{subfigure}
	\begin{subfigure}{0.33\textwidth}
		\centering
		\includegraphics[width=1\textwidth]{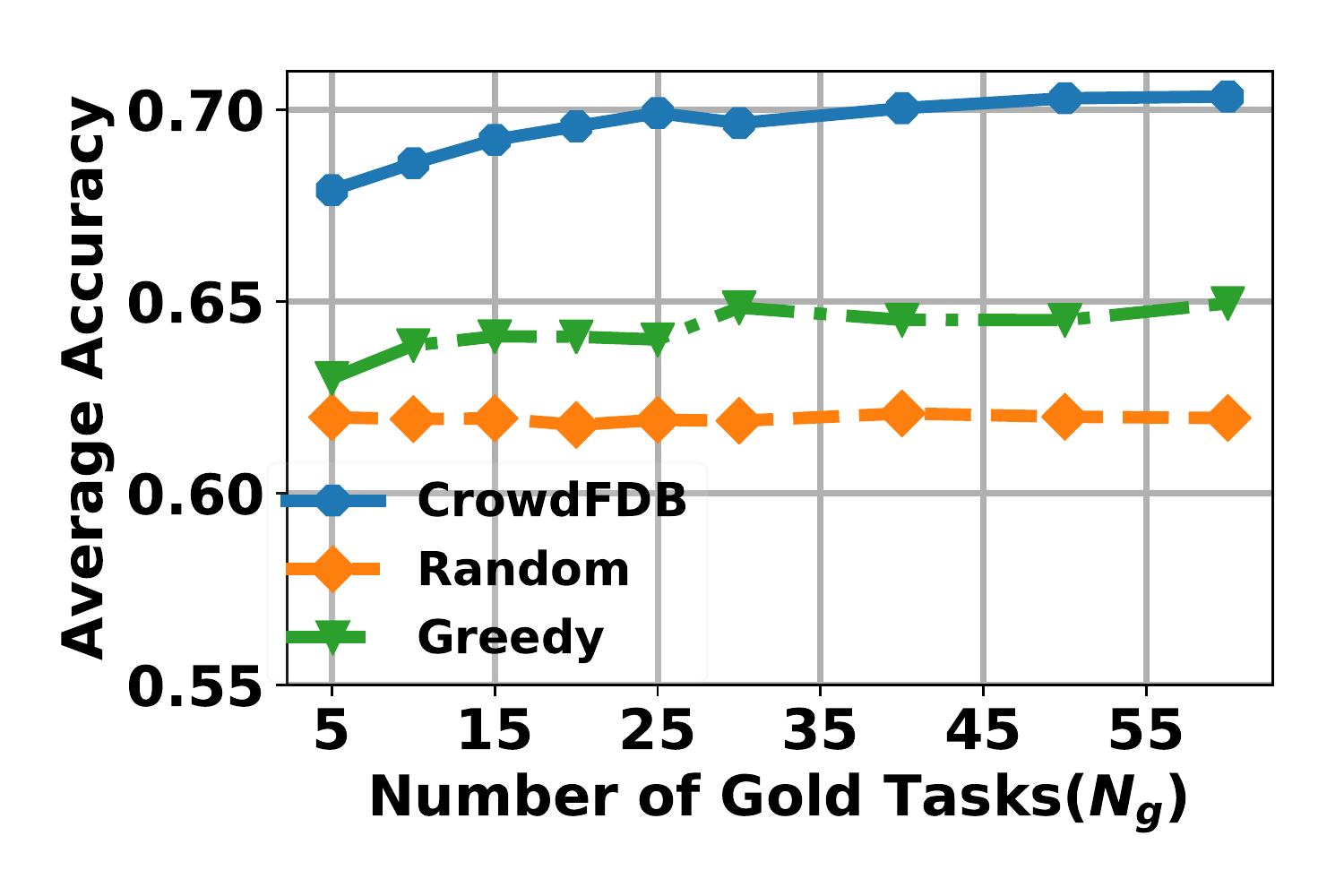}
		\caption{Average Accuracy}
		\label{fig:acc-3_} 
	\end{subfigure}
	\caption{Varying $N_g$ (Number of gold tasks), Settings : Non-Uniform Costs, $\beta = 0.01, \alpha = 0.01, C = 2.5$}
	\label{fig:results-1_}
	\centering
	\begin{subfigure}{0.33\textwidth}
		\centering
		\includegraphics[width=1\textwidth]{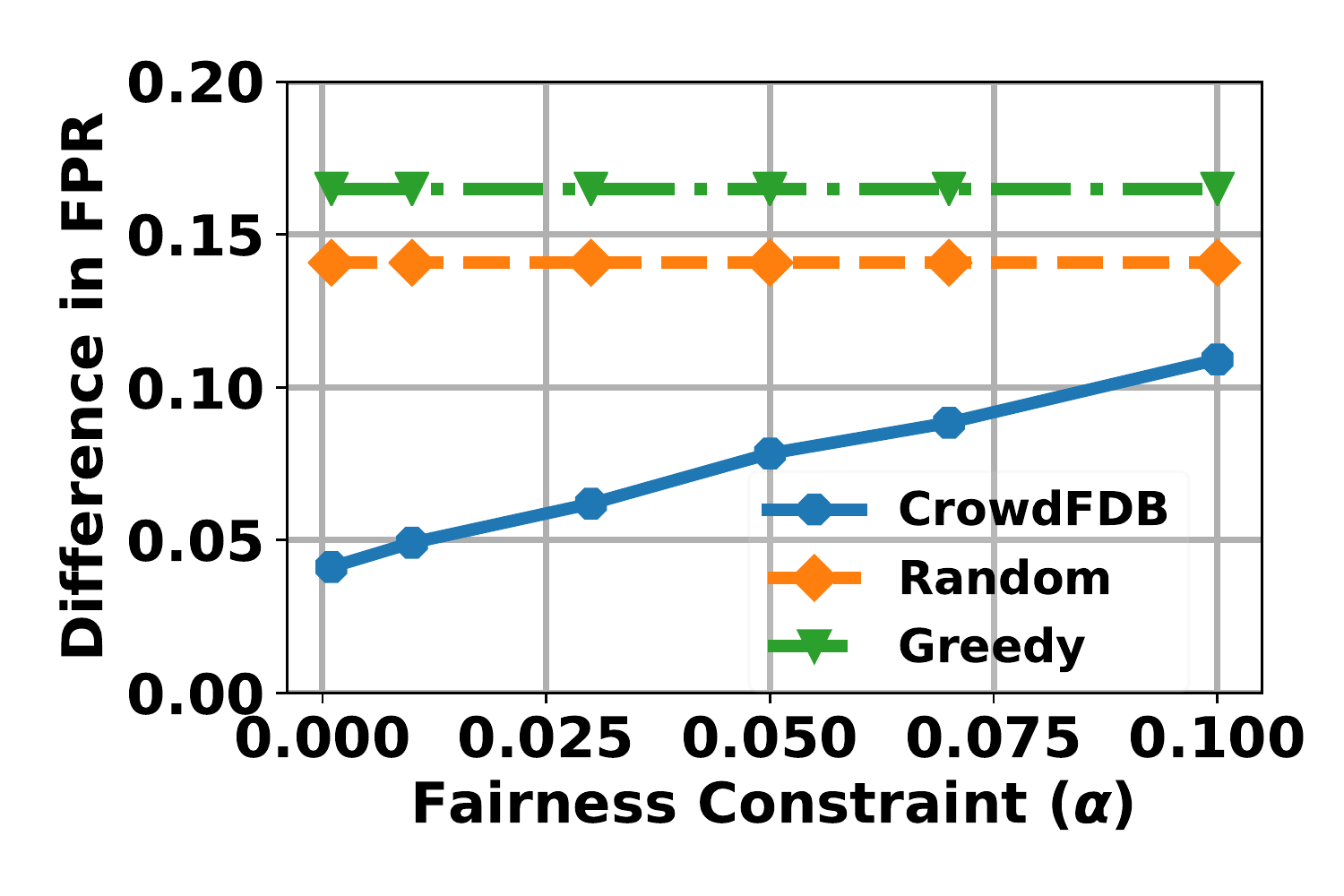}
		\caption{Absolute Difference in FPR}
		\label{fig:fpr-4_} 
	\end{subfigure}
	\begin{subfigure}{0.33\textwidth}
		\centering
		\includegraphics[width=1\textwidth]{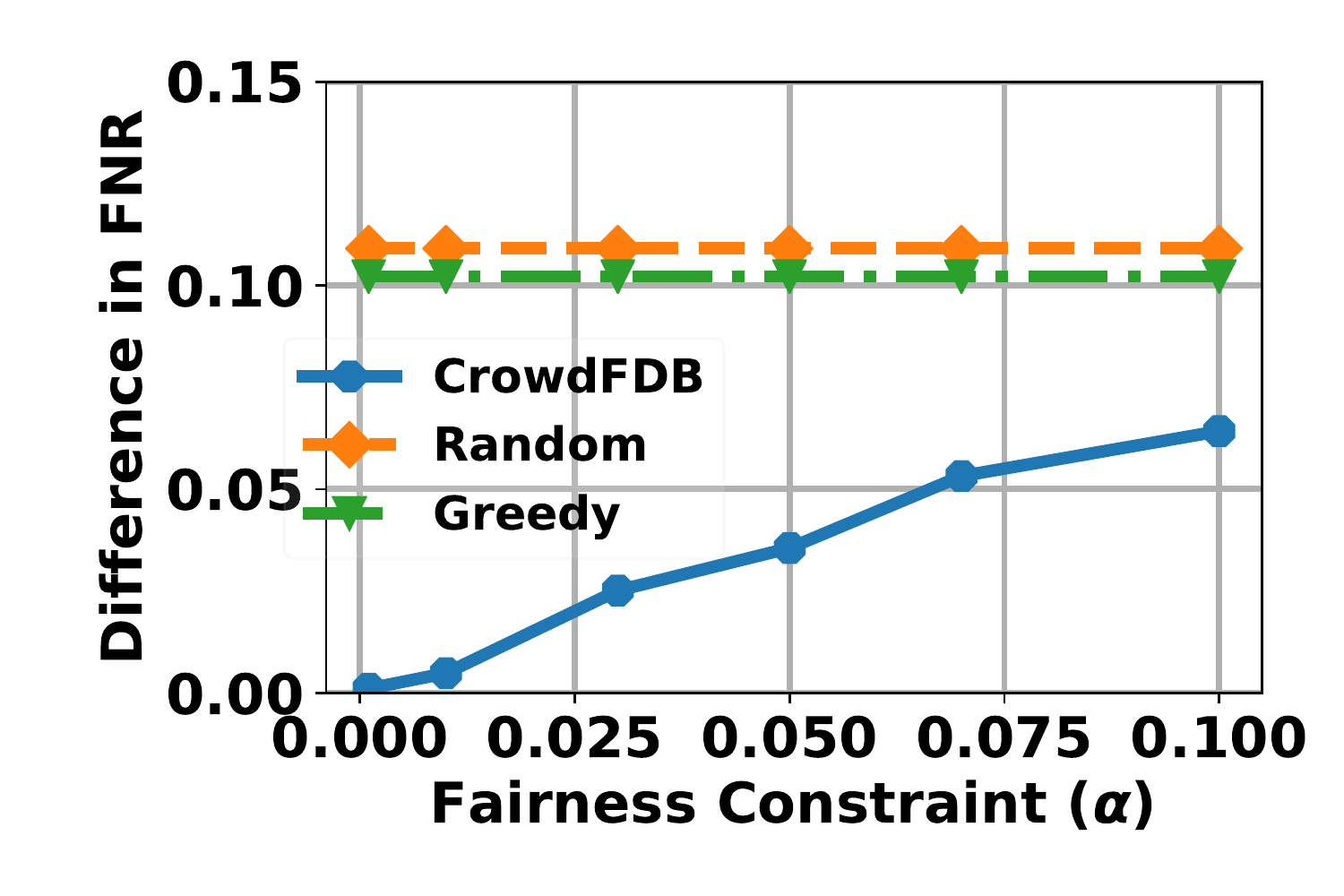}
		\caption{Absolute Difference in FNR}
		\label{fig:fnr-4_} 
	\end{subfigure}
	\begin{subfigure}{0.33\textwidth}
		\centering
		\includegraphics[width=1\textwidth]{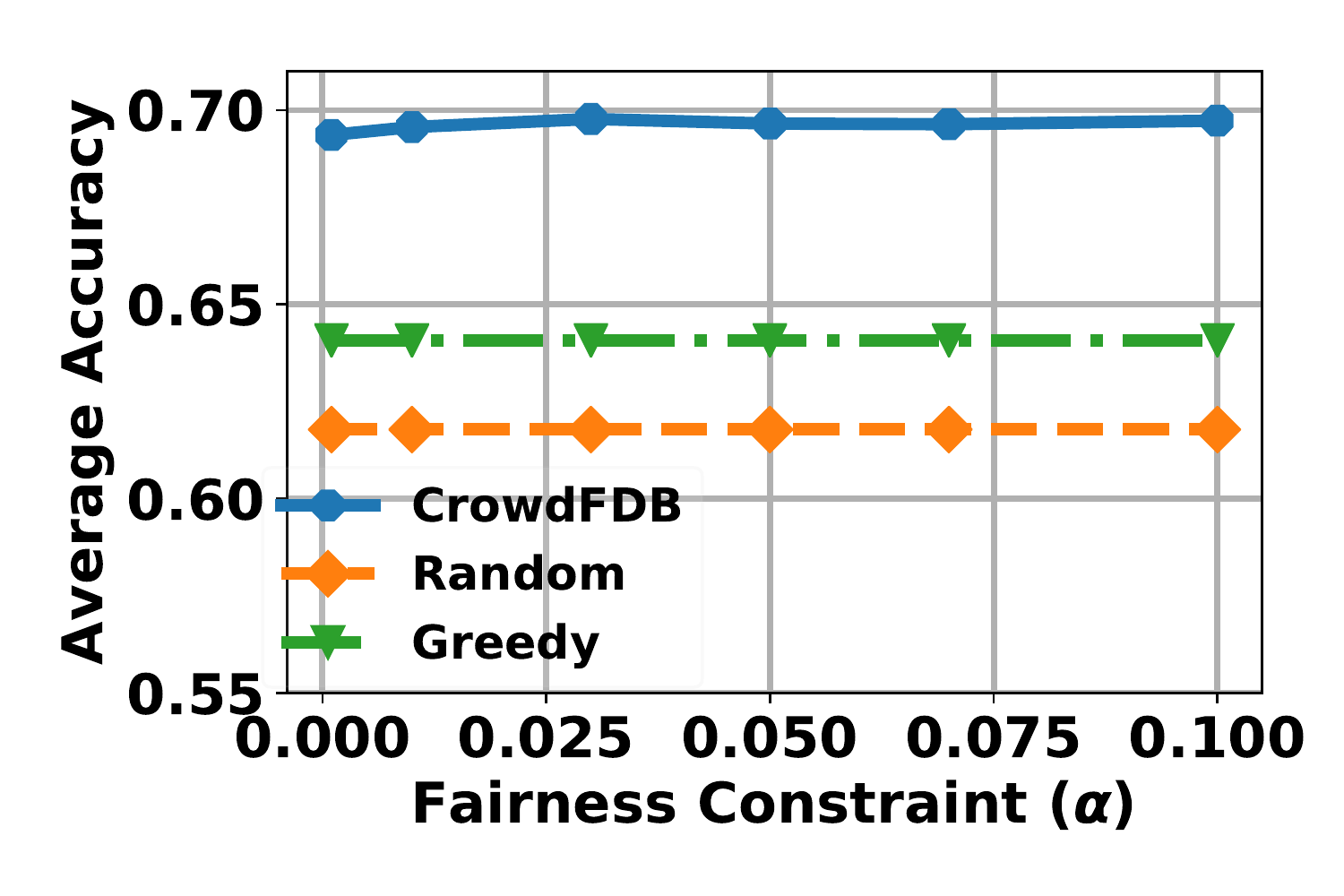}
		\caption{Average Accuracy}
		\label{fig:acc-4_} 
	\end{subfigure}
	\caption{Varying $\alpha$ (Fairness Constraint), Settings : Non-Uniform Costs, $\beta = 0.01, N_g = 20, C = 2.5$}
	\label{fig:results-2_}
\end{figure*}
\begin{figure*}
	\centering
	\begin{subfigure}{0.33\textwidth}
		\centering
		\includegraphics[width=1\textwidth]{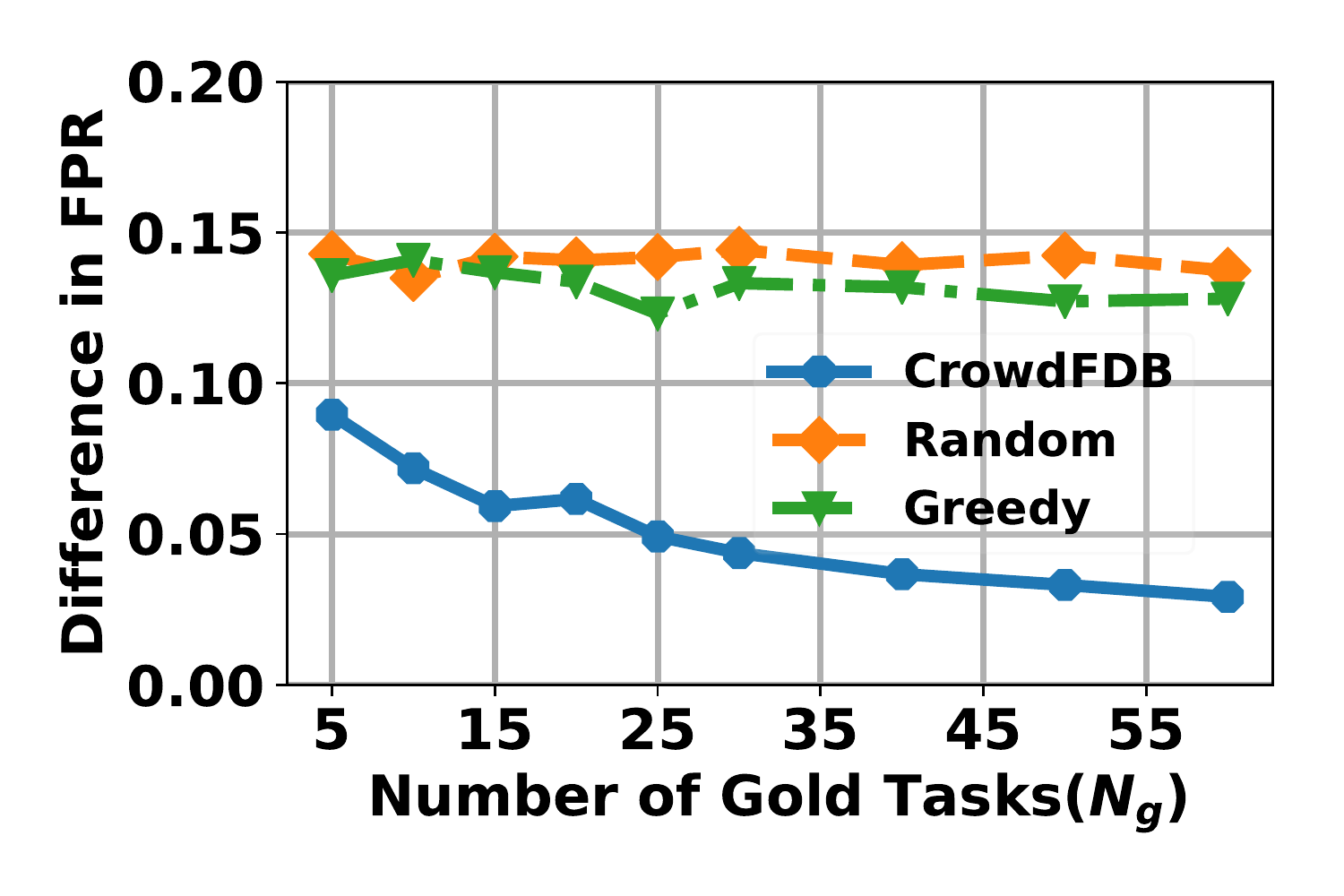}
		\caption{Absolute Difference in FPR}
		\label{fig:_fpr-3_} 
	\end{subfigure}
	\begin{subfigure}{0.33\textwidth}
		\centering
		\includegraphics[width=1\textwidth]{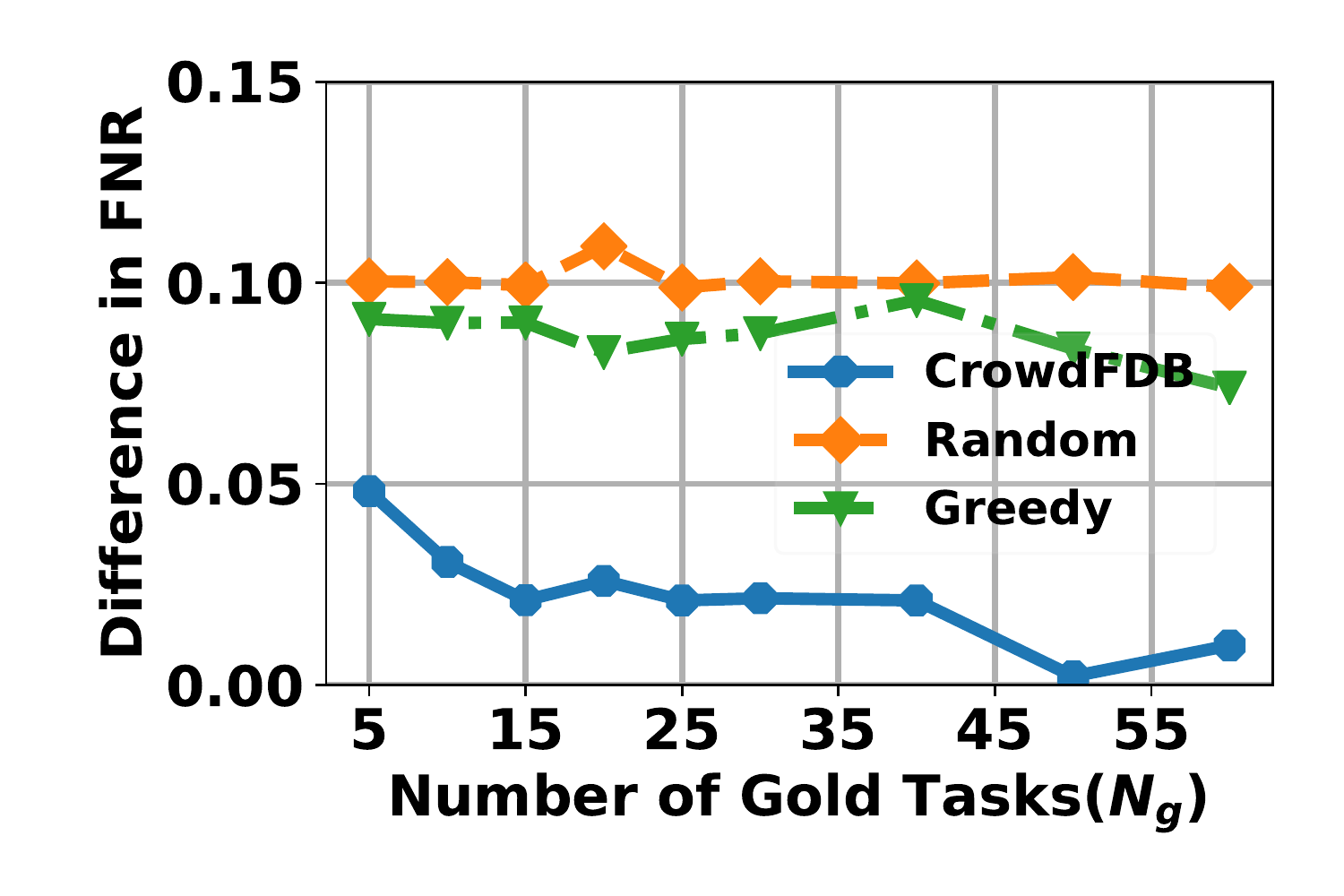}
		\caption{Absolute Difference in FNR}
		\label{fig:_fnr-3_} 
	\end{subfigure}
	\begin{subfigure}{0.33\textwidth}
		\centering
		\includegraphics[width=1\textwidth]{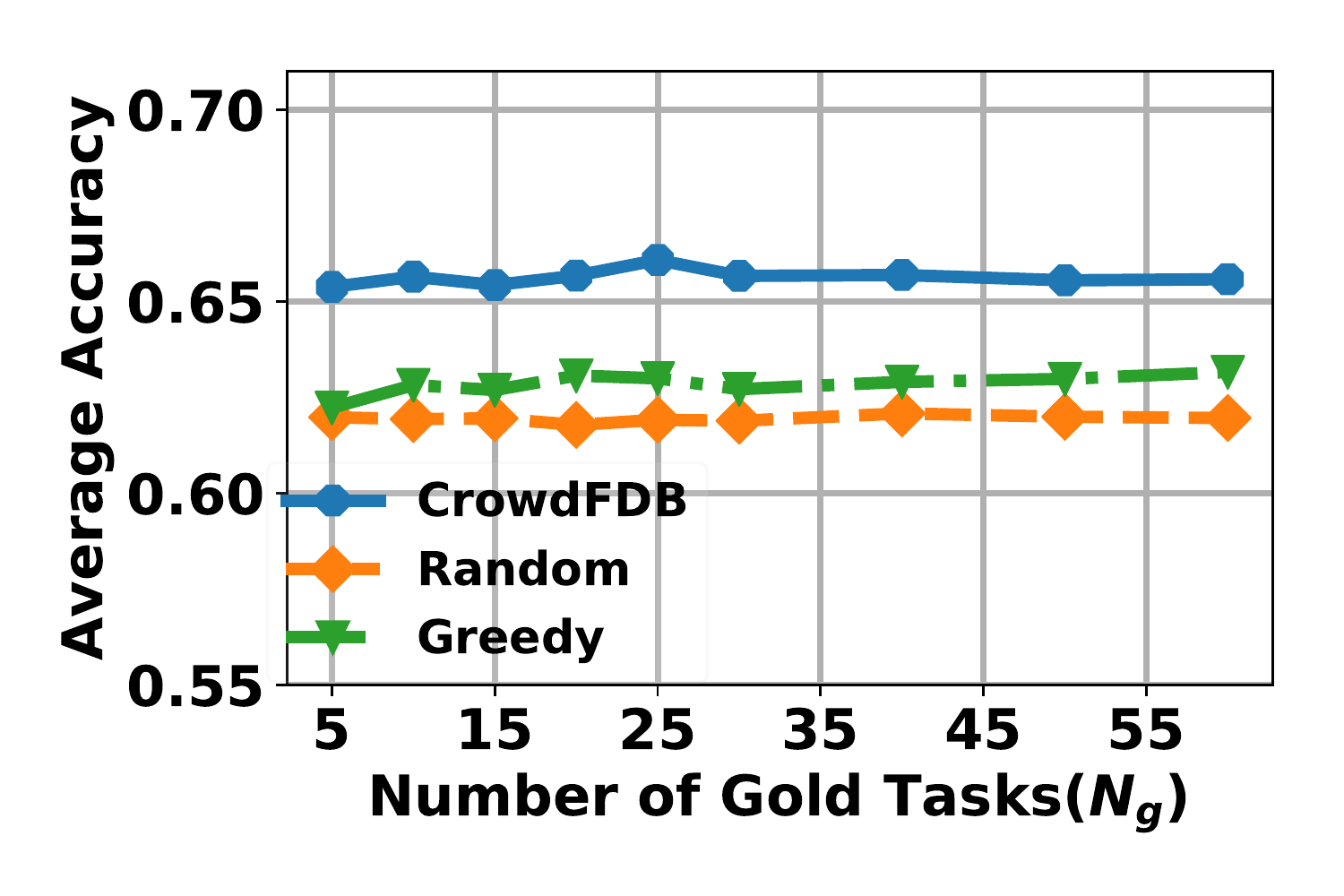}
		\caption{Average Accuracy}
		\label{fig:_acc-3_} 
	\end{subfigure}
	\caption{Varying $N_g$ (Number of gold tasks), Settings : Non-Uniform Costs, $\beta = 0.005, \alpha = 0.01, C = 2.5$}
	\label{fig:results-3_}
	\centering
	\begin{subfigure}{0.33\textwidth}
		\centering
		\includegraphics[width=1\textwidth]{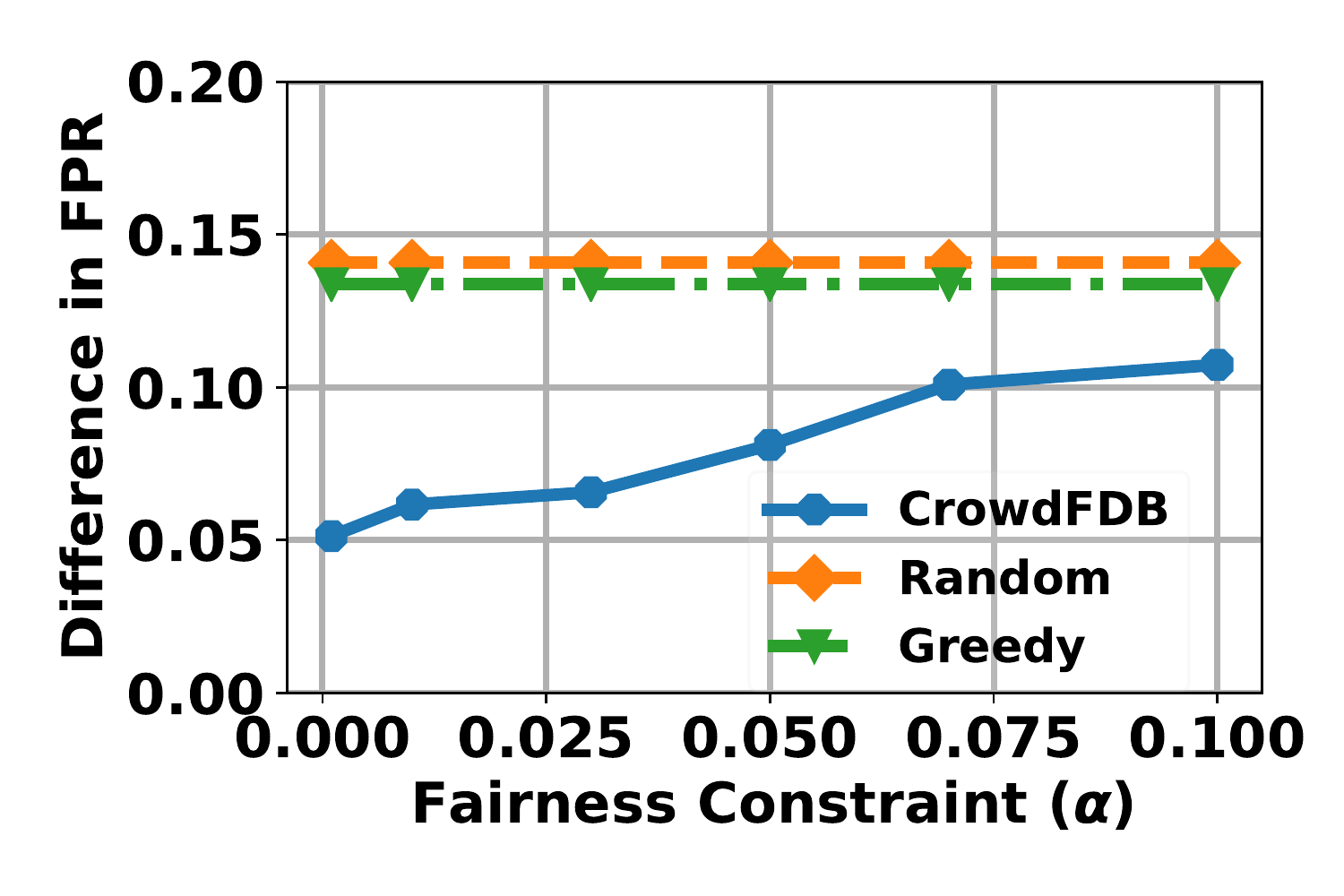}
		\caption{Absolute Difference in FPR}
		\label{fig:_fpr-4_} 
	\end{subfigure}
	\begin{subfigure}{0.33\textwidth}
		\centering
		\includegraphics[width=1\textwidth]{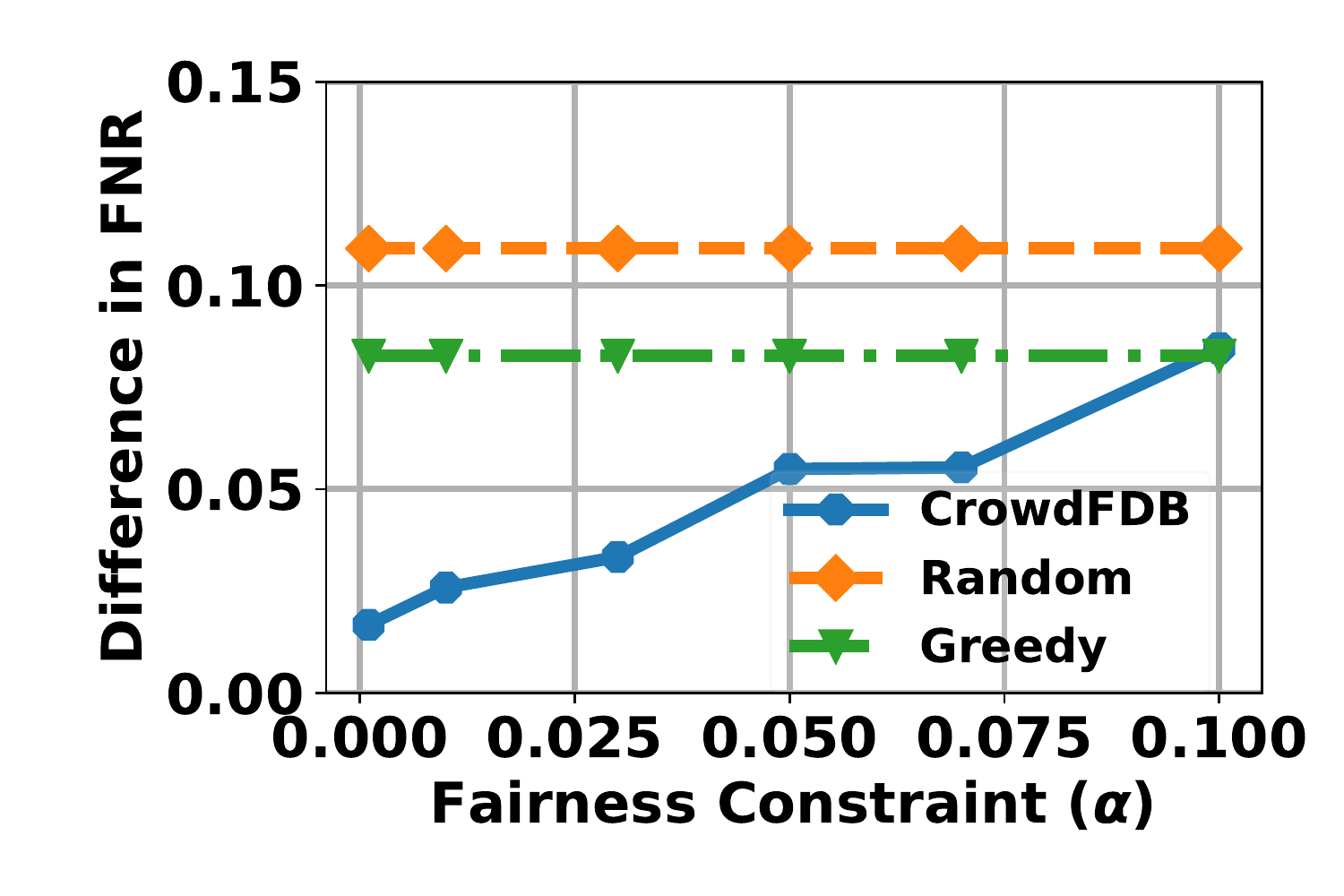}
		\caption{Absolute Difference in FNR}
		\label{fig:_fnr-4_} 
	\end{subfigure}
	\begin{subfigure}{0.33\textwidth}
		\centering
		\includegraphics[width=1\textwidth]{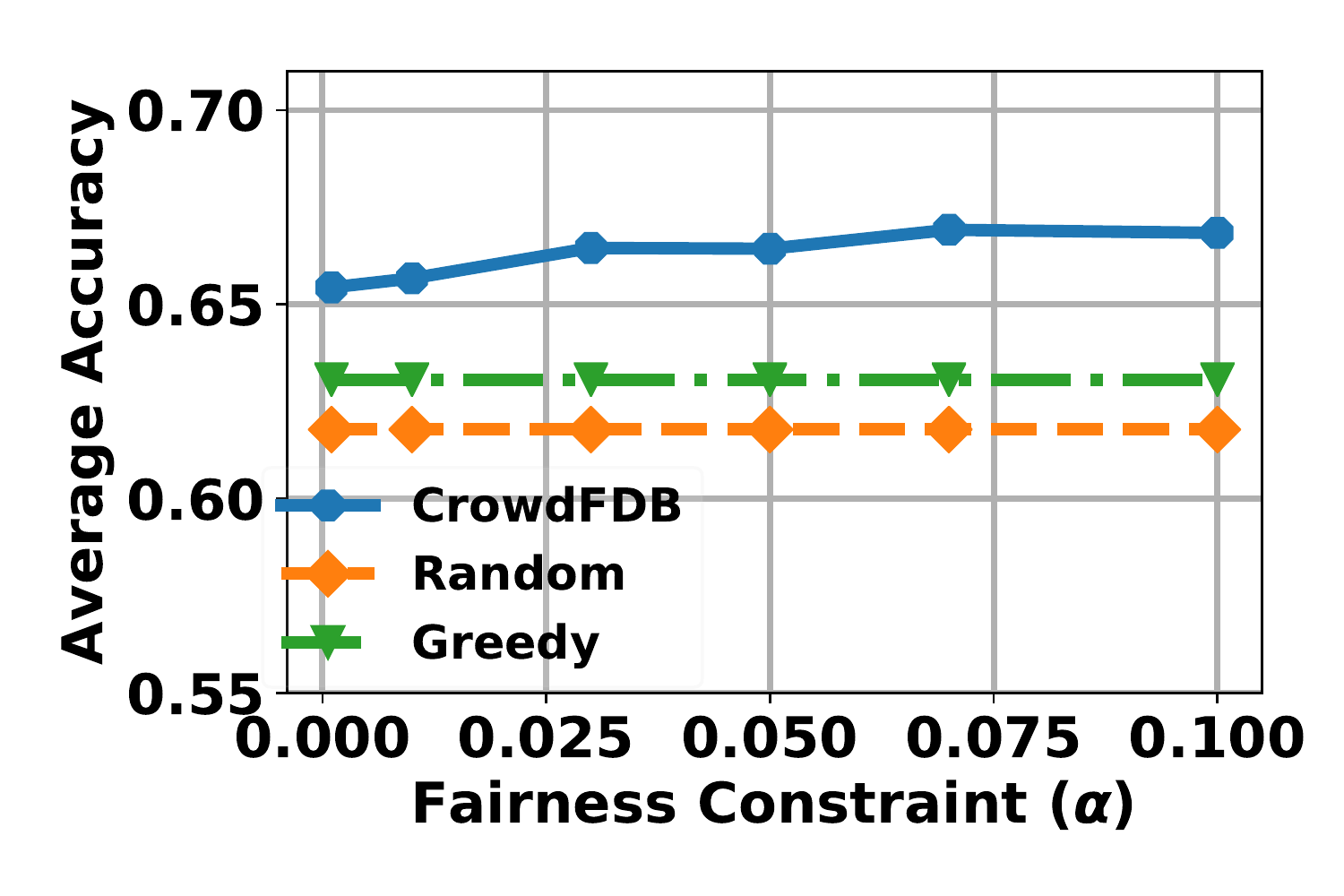}
		\caption{Average Accuracy}
		\label{fig:_acc-4_} 
	\end{subfigure}
	\caption{Varying $\alpha$ (Fairness Constraint), Settings : Non-Uniform Costs, $\beta = 0.005, N_g = 20, C = 2.5$}
	\label{fig:results-4_}
\end{figure*}

\end{document}